\documentclass[10pt,twocolumn,letterpaper]{article}

\usepackage{iccv}              

\definecolor{iccvblue}{rgb}{0.21,0.49,0.74}
\usepackage[pagebackref,breaklinks,colorlinks,allcolors=iccvblue]{hyperref}
\usepackage{url}
\usepackage{booktabs}
\usepackage{multirow}
\usepackage{graphicx}
\usepackage{enumitem}
\usepackage{xcolor}
\usepackage{xspace}
\usepackage{float}
\usepackage{bm}
\usepackage{array} 
\usepackage{algorithm}
\usepackage{algorithmic}
\usepackage{multicol}

\def\paperID{2417} 
\def\confName{ICCV}
\def\confYear{2025}

\title{\name: Probabilistic Integration for Robust and Accurate Point Tracking}

\author{Tingyang Zhang$^{1,2}$ \quad Chen Wang$^{1}$ \quad Zhiyang Dou$^{1,3}$ \quad Qingzhe Gao$^{4}$ \\
Jiahui Lei$^{1}$ \quad Baoquan Chen$^{2}$ \quad Lingjie Liu$^{1}$ \vspace{2pt}\\
$^1$University of Pennsylvania\quad
$^2$Peking University\quad \\
$^3$The University of Hong Kong \quad
$^4$Shandong University \vspace{5pt} \\
{\tt\small \{tyzh,chenw30,zydou,leijh,lingjie.liu\}@seas.upenn.edu;}\\
{\tt\small gaoqingzhe97@gmail.com; baoquan@pku.edu.cn}
}

\newcommand{\name}{ProTracker\xspace}

\begin{document}
\twocolumn[\maketitle
    \centering
    \vspace{-10mm}
    \includegraphics[width=\textwidth]{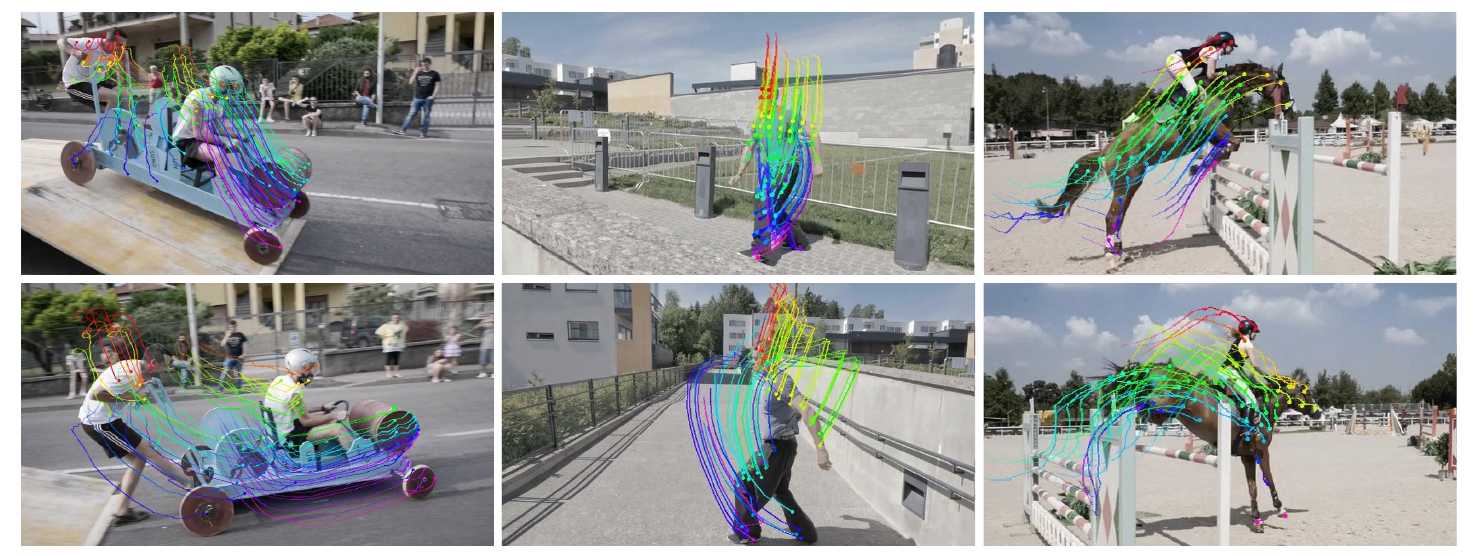}
    \vspace{-6mm}
    \captionof{figure}{Visualization of tracking trajectories in various videos.Our method achieves robust point tracking without suffering from drifting over time, even in challenging scenarios such as occlusions and multiple similar regions.
    }
    \label{fig:teaser}
    \vspace{4mm}
]

\begin{abstract}
We propose \textit{\name}, a novel framework for accurate and robust long-term dense tracking of arbitrary points in videos. Previous methods relying on global cost volumes effectively handle large occlusions and scene changes but lack precision and temporal awareness. In contrast, local iteration-based methods accurately track smoothly transforming scenes but face challenges with occlusions and drift. To address these issues, we propose a probabilistic framework that marries the strengths of both paradigms by leveraging local optical flow for predictions and refined global heatmaps for observations. This design effectively combines global semantic information with temporally aware low-level features, enabling precise and robust long-term tracking of arbitrary points in videos. Extensive experiments demonstrate that \name attains state-of-the-art performance among optimization-based approaches and surpasses supervised feed-forward methods on multiple benchmarks. The code and model will be released after publication.

\end{abstract}
\vspace{-5mm}    
\vspace{-10pt}
\section{Introduction}
Point tracking models~\cite{1641022,longrangemotion,lowe2004sift,rublee2011orb,doersch2022tap,karaev2023cotracker} provide critical motion and deformation cues in scenes, thus they are essential for video analysis, especially for tasks like 4D reconstruction~\cite{lei2024moscadynamicgaussianfusion, stearns2024dynamicgaussianmarblesnovel,som2024} and video editing~\cite{Gu_2024_CVPR}. The recent focus of point tracking is long-term dense tracking of any pixel in a video, also known as Tracking Any Point~(TAP)~\cite{doersch2022tap}.
Existing methods can be broadly classified into two categories. 1) \textit{Supervised tracking models}~\cite{harley2022particle,karaev2023cotracker,xiao2024spatialtracker,doersch2023tapir,cho2024local,cho2024flowtrack,taptr,doersch2022tap}. Specifically, TAP-net~\cite{doersch2022tap} predicts trajectories by generating heatmaps that capture the relationship between the target point and the rest of the frames, while some others~\cite{harley2022particle,karaev2023cotracker,xiao2024spatialtracker,taptr} iteratively refine the trajectory of the same point within a temporal window. 
These supervised learning-based trackers have achieved promising results on existing benchmarks, but they often struggle to generalize to out-of-domain inputs, as they are typically trained on specific datasets. Some of them either disregard temporal information~\cite{doersch2022tap} or suffer
from context drift and loss particularly during extended occlusions as they rely on sliding window techniques~\cite{karaev2023cotracker,xiao2024spatialtracker,harley2022particle,taptr}. 
2) \textit{Optimization based models}~\cite{wang2023tracking,song2024track,li2024decomposition,dino_tracker_2024}. Based on test-time optimization, they have gained attention by leveraging the priors in foundation models trained on web-scale datasets. For instance, some methods~\cite{wang2023tracking,song2024track,li2024decomposition} represent the entire scene as a quasi-3D canonical volume and use 3D bijections to map local coordinates to a global 3D canonical space, allowing for consistent tracking of points.
However, the proxy canonical space represented by neural networks tends to be overly smooth, which limits tracking accuracy. DINO-Tracker~\cite{dino_tracker_2024} fine-tunes a feature extractor and heatmap refiner using the strong semantic priors from DINOv2 to track through long-term occlusions. However, challenges arise when the features are not distinct enough or when multiple similar parts are present in the scene.


In this paper, we present \name for accurate and robust point tracking. The key idea of our method is a bidirectional Probabilistic Integration for both optical flow predictions and long-term correspondences, inspired by Kalman Filter~\cite{kalman1960new}.
Specifically, we begin with removing incorrect initial predictions to reduce their negative impact on subsequent estimations with a hybrid filter including an object-level filter~\cite{ravi2024sam2} and a geometry-aware feature filter~\cite{zhang2024telling}. 
For the remaining rough optical flow predictions, we address the inherent noise in optical flow estimates by introducing a probabilistic integration method that treats each prediction as a Gaussian distribution and merges them into a single Gaussian distribution to identify the most likely point prediction.
The integration is done in both forward and backward directions for highly accurate and robust flow estimation. 
However, optical flow is limited to visible objects and tends to fail when a point disappears and then reappears in a different location, resulting in missing segments in the trajectory. 
To improve performance in challenging long-term point tracking as well as the occlusion problem, we train a long-term feature correspondence model and use it to identify keypoint positions across frames with discriminative features. Then, we jointly integrate flow estimation and long-term keypoints to obtain the final prediction. 
This combination equips the model to robustly recover trajectory segments and mitigate drift during long-term tracking.


We conduct extensive experiments to evaluate our method on TAP-Vid benchmarks. Among optimization-based approaches, our method surpasses all previous methods across all metrics. Additionally, it demonstrates competitive performance even when compared to supervised feed-forward methods and achieves the highest accuracy in position estimation among all approaches.

In summary, our contributions are as follows:
\begin{itemize}
   \item We propose \name, a novel probabilistic integration framework that merges multiple rough predictions and significantly enhances the accuracy and robustness of point tracking. 
   \item We incorporate long-term correspondence matching into our probabilistic integration framework to address both long-term tracking and occlusion, enabling precise point tracking over extended durations.
   \item Our method achieves the state-of-the-art performance among test-time training approaches while demonstrating competitive results compared to data-driven methods.

\end{itemize}

\section{Related Work}
\noindent\textbf{Optical flow} aims to establish dense motion estimations between consecutive frames. Classical methods ~\cite{horn1981determining, lucas1981iterative, brox2009large, brox2004high} optimize warp field with smoothness as regularization. Modern data-driven methods ~\cite{ilg2017flownet,dosovitskiy2015flownet,huang2022flowformer,sun2018pwc,teed2020raft} learn deep neural networks to generate or refine flow predictions based on large amounts of annotated data, which has significantly improved performance. Although optical flow methods can accurately predict displacements between adjacent frames, they often fail when the displacement is too large due to biases in the training data, tending to keep points stationary. This makes optical flow unsuitable for direct long-term tracking. Even chaining flow predictions across frames can lead to drift and other issues. In our approach, we use RAFT~\cite{teed2020raft} as the primary tracking tool, with the aid of additional models to perform precise point tracking.

\vspace{3pt}
\noindent\textbf{Dense correspondence} involves finding pixel-level matches between an arbitrary pair of images. Correlation volumes are constructed to measure the similarity between pairs of pixels based on classic~\cite{lowe2004sift} or learning-based~\cite{cho2021cats,melekhov2019dgc,rocco2020efficient,truong2020glu} feature descriptor, and the accurate point matches are decided accordingly. Recently, large pretrained visual foundation models~\cite{radford2021learning,caron2021emerging,oquab2023dinov2,rombach2022high} have shown their ability to extract powerful features and can be combined for robust matching across different scene/object appearances~\cite{hedlin2024unsupervised,luo2024diffusion,tang2023emergent,zhang2024telling,cheng2024zero}. While directly using these correspondences for point tracking lacks accuracy~\cite{aydemir2024visualfoundationmodelsachieve,dino_tracker_2024} due to the lower resolution of the features compared to the original image, they can effectively serve as a filtering tool to discard incorrect predictions.
\begin{figure*}[!t]
    \centering
    
    \vspace{-20pt}
    \includegraphics[width=\textwidth]{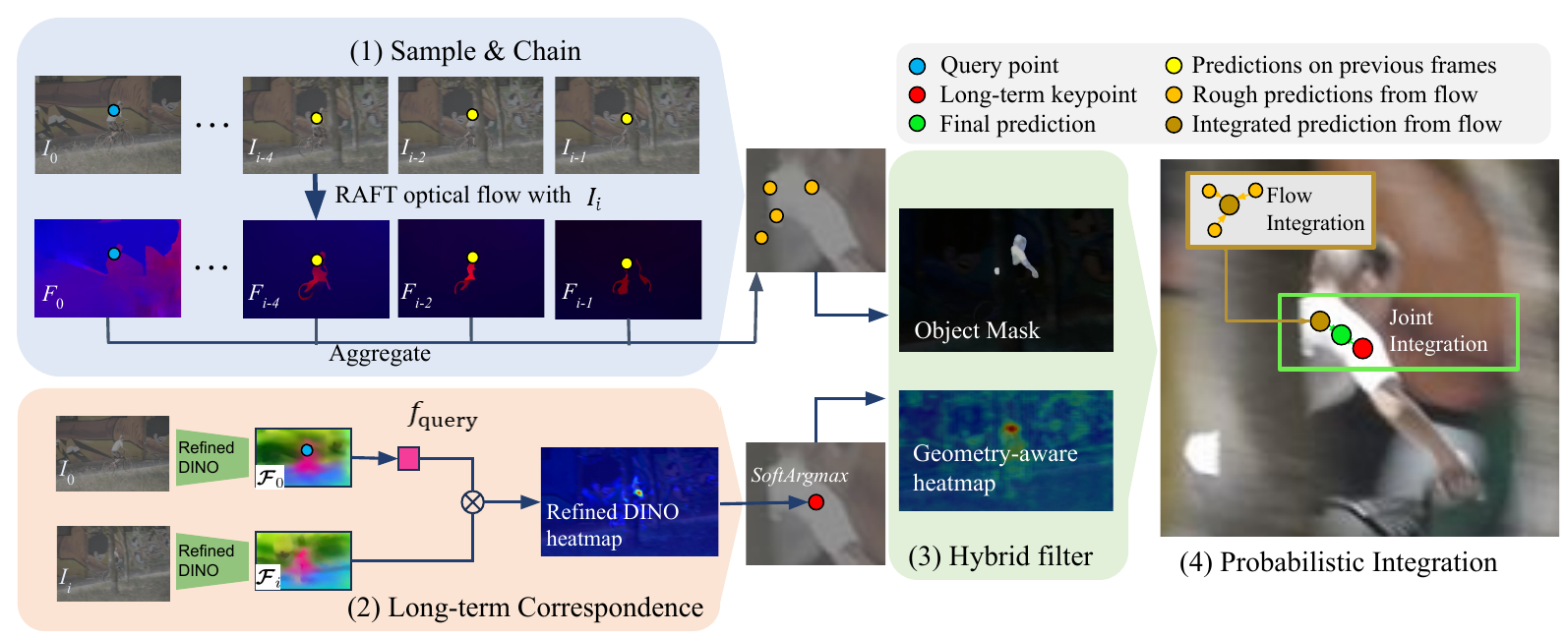}
    \vspace{-20pt}
    \caption{Pipeline overview of our proposed method. (1) Sample \& Chain: Key points are initially sampled and linked through optical flow chaining to produce preliminary trajectory predictions. (2) Long-term Correspondence: Key points are re-localized over longer time spans to maintain continuity, even for points that temporarily disappear.  (3) Hybrid Filter: Masks and feature filters are applied to remove incorrect predictions, reducing noise for subsequent steps. (4) Probabilistic Integration: Filtered flow predictions across frames are first integrated and then combined with long-term keypoint to produce the final prediction, producing smoother and more consistent trajectories. } 
    \vspace{-15pt}
    \label{fig:pipeline}
\end{figure*}

\vspace{3pt}
\noindent\textbf{Tracking any point} aims to track arbitrary points across a whole video, recovering the full trajectory and occlusion state. TAP-net~\cite{doersch2022tap} directly predicts via finding the target in a refined heatmap. PIPs~\cite{harley2022particle} proposes to iteratively refine the trajectory within a temporal window according to the spatial context. Many attempts have been made to improve the refinement process. Co-tracker~\cite{karaev2023cotracker} counted in the relation between points and designed a self-attention to support them with each other. SpatialTracker~\cite{xiao2024spatialtracker} lifts points to 3d space and performs tracking with spatially meaningful information. TAPTR~\cite{taptr} treats points as queries and updates them in a DETR~\cite{carion2020endtoendobjectdetectiontransformers} style. Some other methods like TAPIR~\cite{doersch2023tapir} and LocoTrack ~\cite{cho2024local} adopt a coarse-to-fine strategy, dividing the tracking process into initialization and iterative optimization phases, which allows the well-initialized points to guide the trajectory in other frames. While those supervised methods may be limited to their spatial or temporal field of view due to large memory cost, Omnimotion~\cite{wang2023tracking} first proposes to learn a 3d representation for each video with color and pre-computed optical flow as self-supervision, in which a bijective mapping enables the query of any point in a different frame. Decomotion~\cite{li2024decomposition} decomposes the scene representation into static and dynamic and utilizes a temporal invariant feature as extra supervision. CaDeX++~\cite{song2024track} leverages a depth estimator to speed up and a more efficient deformation network. DINO-Tracker~\cite{dino_tracker_2024} trains a delta feature extractor as compensation for the powerful DINO~\cite{oquab2023dinov2} feature. MFT~\cite{neoral2024mft} is a zero-shot method that directly chains optical flow and selects the most reliable estimation as the final tracking prediction. However, problems like drift may occur when facing long videos.


\section{Method}

Given an image sequence $\{{I^t}\}_{t=1}^T$ from a monocular video, our goal is to take a query pixel $\bm{p}^t \in \mathbb{R}^2$ from an arbitrary frame $I^t$ as input and predict its trajectories $\{\hat{\bm{p}}^t\}_{t=1}^T$ over the video, along with the occlusion prediction $\{\hat{o}^t\}_{t=1}^T$, which is known as the TAP (Tracking Any Point) problem. 


As shown in Fig.~\ref{fig:pipeline}, our pipeline first obtains both initial rough optical flow predictions from multiple previous frames and long-term correspondence predictions. We filter unreliable point predictions with an object-level segmentation model~\cite{ravi2024sam2} and geometry-aware semantic features~\cite{zhang2024telling} (Sec.~\ref{sec:object_segmentation}). Then, we integrate multiple optical flow predictions in a probabilistic manner to get an integrated prediction from flow (Sec.~\ref{sec:kalman_filtering}). We further use a joint probabilistic integration between optical flow prediction and long-term semantic correspondence prediction to aggregate both global and local contexts, prevent drift and allow re-localization after reappearance (Sec.~\ref{sec:correspondence_tracking}).

\begin{figure*}[t]
    \centering
    \vspace{-20pt}
    
    \includegraphics[width=\textwidth]{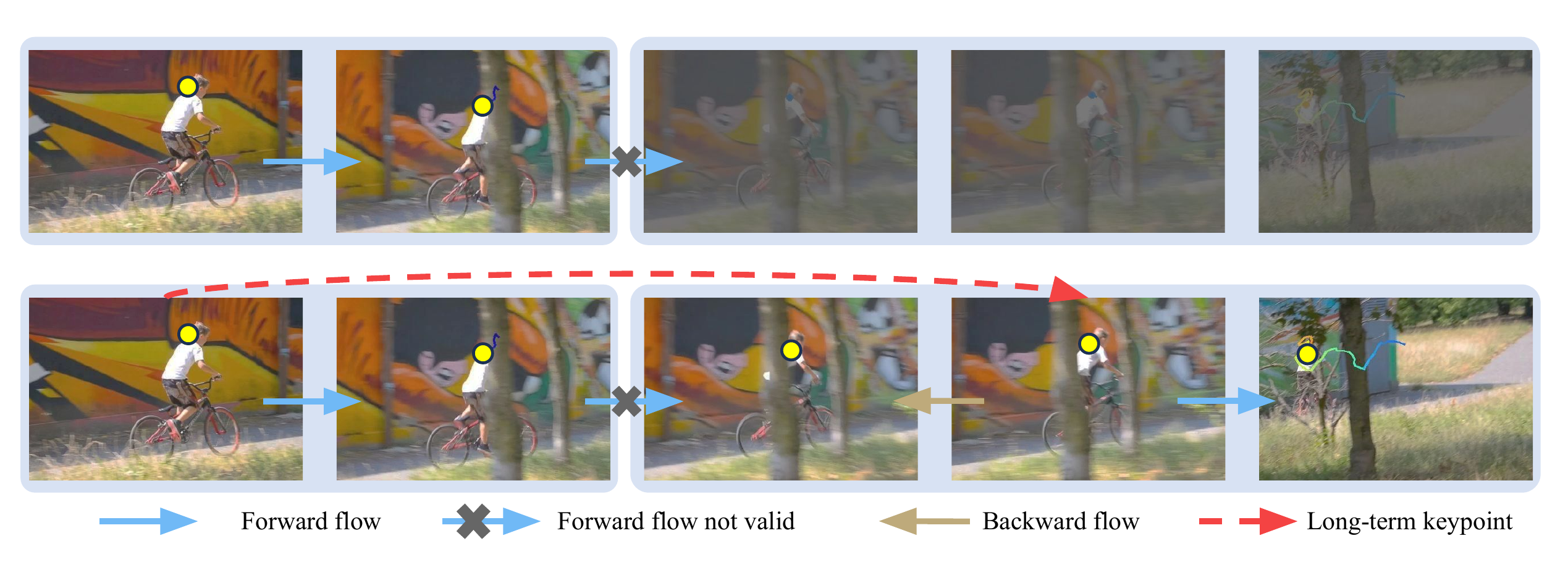}

    \vspace{-20pt}
    \caption{
    Bidirectional Probabilistic Flow Integration. Top row: Optical flow effectively tracks a point in the short term but may fail under occlusion due to its local nature. Bottom row: Long-term correspondence aids in globally relocating the target when the tracked point reappears. Once relocation is achieved, optical flow can resume tracking in the surrounding frames.
    }
    
    \label{fig:empty_flow}
    \vspace{-15pt}
\end{figure*}

\subsection{Hybrid Filter}
\label{sec:object_segmentation}
Since our method relies on rough predictions from optical flow and long-term correspondence for probabilistic integration (as shown in Fig.~\ref{fig:pipeline}), inaccurate rough predictions can lead to cumulative errors and distort entire trajectories, which may significantly degrade tracking accuracy. To mitigate these issues, we propose a hybrid filter to abandon these predictions and avoid using them in the following probabilistic integration.

Our hybrid filter consists of an object-level filter and a geometry-aware feature filter.
First, an object-level segmentation model~\cite{ravi2024sam2} generates masks associated with target points, filtering out predictions outside relevant objects and using global context to mitigate the impact of outliers; see ablation study in Fig.\ref{fig:ablation}. This step significantly benefits optical flow-based systems like RAFT~\cite{teed2020raft}, which often struggle with occlusions due to their reliance on local feature matching.

To further reduce ambiguity between semantically similar points and prevent flickering across different regions, an additional geometry-aware feature extractor~\cite{zhang2024telling} is employed. For each point, if its feature similarity to the original query point falls below 0.5, the point is classified as occluded, ensuring that only reliable predictions are retained and preventing errors from propagating due to semantic ambiguity. Together, the object-level and geometry-aware filters consist of a module which can distinguish different points from a global semantic perspective, thereby avoiding confusion between different objects or similar regions caused by local inductive bias from optical flow.


\subsection{Bidirectional Probabilistic Flow Integration }
\label{sec:kalman_filtering}
We introduce a probabilistic integration strategy inspired by the Kalman filter~\cite{kalman1960new}, enabling frame-by-frame trajectory recovery from both semantics and low-level features. Our method incorporates both forward and backward passes, leveraging forward and backward flows to reconstruct complete trajectories. Further details are provided below.


\paragraph{Forward Integration} 
Our method sequentially predicts point trajectory and occlusion on the current frame based on previous predictions. Since every estimate is subject to noise, we extend both the track predictions and the optical flow into two-dimensional Gaussian distributions. 
Thus, we denote the predictions of frame $i$ as $(\bm{\mu}_i, \bm{\Sigma}_{i})$, where $\bm{\mu}_i$ and $\bm{\Sigma}_{i}$ are the mean and covariance of the Gaussian distribution.
We further assume these Gaussian distributions are isotropic and simplify the covariance matrix as $\bm{\Sigma}_i = \sigma_i^2 \bm{I}$, where $\bm{I}$ is the identity matrix. 
For the initial frame, we assume zero uncertainty ($\sigma_0 = 0$). For any frame $i > 0$, we first calculate the flow chain estimations based on previous frames $\{j_1,j_2,...,j_n\}$. Given the prediction for frame $j \in\{j_1,j_2,...,j_n\}\:(j<i)$, denoted as \((\bm{\mu}_j, \sigma_j)\), and the optical flow from frame \(j\) to frame \(i\) denoted as \((\bm{f}_{ji}, \sigma_{ji})\), we can obtain the mean and variance of the prediction for frame \(i\) after the filtering process in Sec.~\ref{sec:object_segmentation}:
\begin{align}
    \bm{\mu}_{ji} &= \bm{\mu}_j + \bm{f}_{ji}, \\ (\sigma_{ji}^2\bm{I}) &= \bm{J}_{f_{ji}} (\sigma_{j}^2\bm{I}) \bm{J}_{f_{ji}}^T + (\sigma_{f_{ji}}^2\bm{I}),
\end{align}
where $\bm{\mu}_{ji}$ and $\sigma_{ji}$ are the mean and variance of the chained prediction from frame $j$ to frame $i$. \(\bm{J}_{f_{ji}}\) is the Jacobian matrix of the flow \(f_{ji}\) with respect to the position \(\mu_j\), and \(\sigma_{f_{ji}}\) is the variance of the optical flow. For ease of computation, we assume \(\bm{J}_{f_{ji}}\) to be orthogonal, then:
\begin{equation}
\sigma_{ji}^2 = \sigma_j^2 + \sigma_{f_{ji}}^2.
\end{equation}

We then combine the predictions from previous frames $j\in\{j_1,j_2,...,j_n\}$ to frame $i$ by assuming that they are independent. Since the product of the probability density function (PDF) of Gaussian distributions remains a Gaussian, we can merge multiple predictions for frame \(i\) from different previous frames, \(\{\bm{\mu}_j\}\) with their corresponding variances \(\{\sigma_j^2\}\), into a single refined estimate. The refined mean is computed as a weighted linear combination of \(\{\bm{\mu}_j\}\), with the weights determined by the inverse of their variances:

\begin{equation}
\bm{\mu}_i = \frac{\sum_j \bm{\mu}_{ji} / \sigma_{ji}^2}{\sum_j 1 / \sigma_{ji}^2}.
\end{equation}

Similarly, the refined variance is updated according to the following formula:
\begin{equation}
\quad \sigma_i^2 = \frac{1}{\sum_j 1 / \sigma_{ji}^2}.
\end{equation}

However, previous predictions are typically correlated. To account for these correlations and simplify the calculations, we introduce a constant correlation coefficient \(\textit{p}\) between any pair of estimates from previous frames. The final refined estimates are then given by:
\begin{equation}
\bm{\mu}_i^{f} = \frac{\sum_j \bm{\mu}_{ji} / \sigma_{ji}^2}{\sum_j 1 / \sigma_{ji}^2}, \quad \sigma_i^{f} = \sqrt{\frac{(N-1) \cdot p + 1}{\sum_j 1 / \sigma_{ji}^2}}.
\end{equation}
where \(\bm{\mu}_i\) represents the final predicted position for frame \(i\), and \(\sigma_i^2\) is the combined variance, reflecting the confidence in the estimate based on multiple sources. Following MFT~\cite{neoral2024mft}, we adopt \(\{\infty, 1, 2, 4, 8, 16, 32\}\) as the time intervals, meaning that each frame's prediction is computed by combining results from these previous frames, if the target point is predicted visible in those frames. Here, \(\infty\) refers to the first frame of the video. If all predictions from previous frames to current frame \textit{i} are invalid, the point is marked \textbf{occluded} in frame \textit{i}. Once we reach the last frame and complete the forward tracking, we perform a similar backward pass to recover points that might have been missed, as some points are more easily tracked from future frames because the optical flow from previous frames is no longer accurate due to long-term occlusion.

\paragraph{Backward Integration} 
After the forward pass, we run a backward pass starting from the last frame, focusing on points previously marked as occluded. For any frame \textit{i}, given the prediction for frame $j\in\{j_1,j_2...j_n\}\:(j>i)$, denoted as \((\bm{\mu}_j, \sigma_j)\), and the optical flow from frame \(j\) to frame \(i\), \((f_{ji}, \sigma_{ji})\), we can obtain the mean and variance of the prediction for frame \(i\) after the same filtering process:

\begin{equation}
\bm{\mu}_i^{b} = \frac{\sum_j \bm{\mu}_{ji} / \sigma_{ji}^2}{\sum_j 1 / \sigma_{ji}^2}, \quad \sigma_i^{b} = \sqrt{\frac{(N-1) \cdot p + 1}{\sum_j 1 / \sigma_{ji}^2}}.
\end{equation}

If a point marked as occluded in the forward pass is visible in the backward pass, we adopt the backward result instead. Otherwise, we retain the forward prediction. This ensures less visible point ignored by the tracker, particularly in cases of occlusion. The backward pass helps recover points that are difficult to track from earlier frames but can be more easily tracked from later ones.

\begin{table*}
    \centering
    \resizebox{1.\textwidth}{!}{
    \begin{small}
\begin{tabular}{p{2cm}>{\centering\arraybackslash}p{0.9cm}>{\centering\arraybackslash}p{0.9cm}>{\centering\arraybackslash}p{0.9cm}>{\centering\arraybackslash}p{0.9cm}>{\centering\arraybackslash}p{0.9cm}>{\centering\arraybackslash}p{0.9cm}>{\centering\arraybackslash}p{0.9cm}>
{\centering\arraybackslash}p{0.9cm}>{\centering\arraybackslash}p{0.9cm}>{\centering\arraybackslash}p{0.9cm}>{\centering\arraybackslash}p{0.9cm}}

        \toprule
        \multirow{2}{*}{Method} & \multicolumn{3}{c}{DAVIS-First} & \multicolumn{3}{c}{DAVIS-Strided} & \multicolumn{3}{c}{Kinetics-First} & \multicolumn{2}{c}{BADJA}\\
        \cmidrule(lr){2-4} \cmidrule(lr){5-7} \cmidrule(lr){8-10} \cmidrule(lr){11-12}
        & $\delta_{avg}^x\uparrow$ & OA$ \uparrow$ & AJ$ \uparrow$ & $\delta_{avg}^x\uparrow$ & OA $\uparrow$ & AJ $\uparrow$ & $\delta_{avg}^x \uparrow$ & OA $\uparrow$ & AJ $\uparrow$ &
        $\delta^{seg}\uparrow$ & $\delta^{3px}\uparrow$\\
        
        \cmidrule(lr){1-12}
        Omnimotion~\cite{wang2023tracking}& -    & -    & -    & 67.5 & 85.3 & 51.7 & -    & -    & -  & 45.2 & 6.9  \\
        MFT~\cite{neoral2024mft}       & 66.8 & 77.8 & 47.3 & 70.8 & 86.9 & 56.1 & 60.8 & 75.6 & 39.4 & 50.8 & 7.2\\
        CaDeX++~\cite{song2024track}   & -    & -    & -    & 77.4 & 85.9 & 59.4 & -    & -    & -  & -    & -    \\
        DecoMotion~\cite{li2024decomposition}& 69.9 & 84.2 & 53.0 & 74.4 & 87.2 & 60.2 & -    & -    & -  & -    & -    \\
        DINOTracker~\cite{dino_tracker_2024} & 74.9& 86.4& 58.3 & 78.2 & 87.5 & 62.3 & 69.5 & 86.3 & 55.5 & 72.4 & 14.3\\
        Ours      & \textbf{77.6} & \textbf{87.3} & \textbf{62.0} & \textbf{80.8} & \textbf{88.7} & \textbf{65.3} & \textbf{71.1} & \textbf{89.6} & \textbf{56.7} & \textbf{73.3} & \textbf{14.4}\\
        \bottomrule
    \end{tabular}
    \end{small}
    }
    \vspace{-9pt}
    \caption{
    We compare our method with state-of-the-art \textbf{optimization-based} trackers on the TAP-Vid and BADJA benchmarks. We include MFT, which directly integrates optical flow for tracking. Our method consistently achieves superior performance across all metrics, with the best results \textbf{bolded}.
    }
    \vspace{-10pt}
    \label{tab:performance_unsupervised}
\end{table*}
\begin{table*}
    \centering
    \resizebox{\textwidth}{!}{
    \begin{small}
\begin{tabular}{p{2cm}>{\centering\arraybackslash}p{0.9cm}>{\centering\arraybackslash}p{0.9cm}>{\centering\arraybackslash}p{0.9cm}>{\centering\arraybackslash}p{0.9cm}>{\centering\arraybackslash}p{0.9cm}>{\centering\arraybackslash}p{0.9cm}>{\centering\arraybackslash}p{0.9cm}>{\centering\arraybackslash}p{0.9cm}>{\centering\arraybackslash}p{0.9cm}>
{\centering\arraybackslash}p{0.9cm}>{\centering\arraybackslash}p{0.9cm}}

        \toprule
        \multirow{2}{*}{Method} & \multicolumn{3}{c}{DAVIS-First} & \multicolumn{3}{c}{DAVIS-Strided} & \multicolumn{3}{c}{Kinetics-First} & \multicolumn{2}{c}{BADJA}\\
        \cmidrule(lr){2-4} \cmidrule(lr){5-7} \cmidrule(lr){8-10} \cmidrule(lr){11-12}
        & $\delta_{avg}^x\uparrow$ & OA$ \uparrow$ & AJ$ \uparrow$ & $\delta_{avg}^x\uparrow$ & OA $\uparrow$ & AJ $\uparrow$ & $\delta_{avg}^x \uparrow$ & OA $\uparrow$ & AJ $\uparrow$ &
        $\delta^{seg}\uparrow$ & $\delta^{3px}\uparrow$\\
        \cmidrule(lr){1-12}
        TAP-Net~\cite{doersch2022tap}   & 48.6 & 78.8 & 33.0 & 53.4 & 81.4 & 38.4 & 56.3 & 83.6 & 42.7 & 45.4 & 9.6\\
        PIPs~\cite{harley2022particle}      & 64.8 & 77.7 & 42.2 & 59.4 & 82.1 & 42.0 & 47.6 & 78.5 & 31.1 & 59.6 & 9.4\\
        TAPIR~\cite{doersch2023tapir}     & 70.0 & 86.5 & 56.2 & 74.7 & 89.4 & 62.8 & 63.6 & 86.4 & 52.6 & 68.7 & 10.5\\
        CoTracker~\cite{karaev2023cotracker} & 75.4 & 89.3 & 60.6 & 79.2 & 89.3 & 65.1 & 65.9 & 88.0 & 52.8 & 64.0 & 11.2\\
        SpatialTracker~\cite{xiao2024spatialtracker}& 76.3 & 89.5 & 61.1 & - & -   & -    & 67.1 & 88.3 & 53.9 & 63.6 & 10.6\\
        BootsTAPIR~\cite{doersch2024bootstap}& 74.0 & 88.4 & 61.4 & 78.5 & 90.7   & 66.4    & - & - & - & 72.6 & 13.4\\
        CoTracker3~\cite{karaev24cotracker3}& 76.9 & 91.2 & \textbf{64.4} & - & - & -    & 70.9 & 86.9 & \textbf{57.9} & 71.4 & 11.1\\
        TAPTRv2~\cite{li2024taptrv2} & 75.9 & \textbf{91.4} & 63.5 & 78.8 & \textbf{91.3} & 66.4 & 64.4 & 85.7 & 50.8 & 70.0 & 8.4\\
        LocoTrack~\cite{cho2024local}& 75.3 & 87.2 & 63.0 & 79.6 & 89.9 & \textbf{67.8} & 68.8 & 87.5 & 56.0 &70.2 & 9.9\\
        Ours      & \textbf{77.6} & 87.3 & 62.0 & \textbf{80.8} & 88.7 & 65.3 & \textbf{71.1} & \textbf{89.6} & 56.7  & \textbf{73.3} & \textbf{14.4}\\
        \bottomrule
    \end{tabular}
    \end{small}
    }
    \vspace{-9pt}
    \caption{We compare our method with \textbf{supervised feed-forward} trackers on the TAP-Vid and BADJA benchmarks, where our method achieves the highest \(\delta_{avg}^x\) across all datasets and produces competitive results in OA and AJ. The best results are \textbf{bolded}.}
 \vspace{-18pt}
    \label{tab:performance}
\end{table*}

\begin{figure*}
    \centering
    
    \vspace{15pt}
    \includegraphics[width=\textwidth]{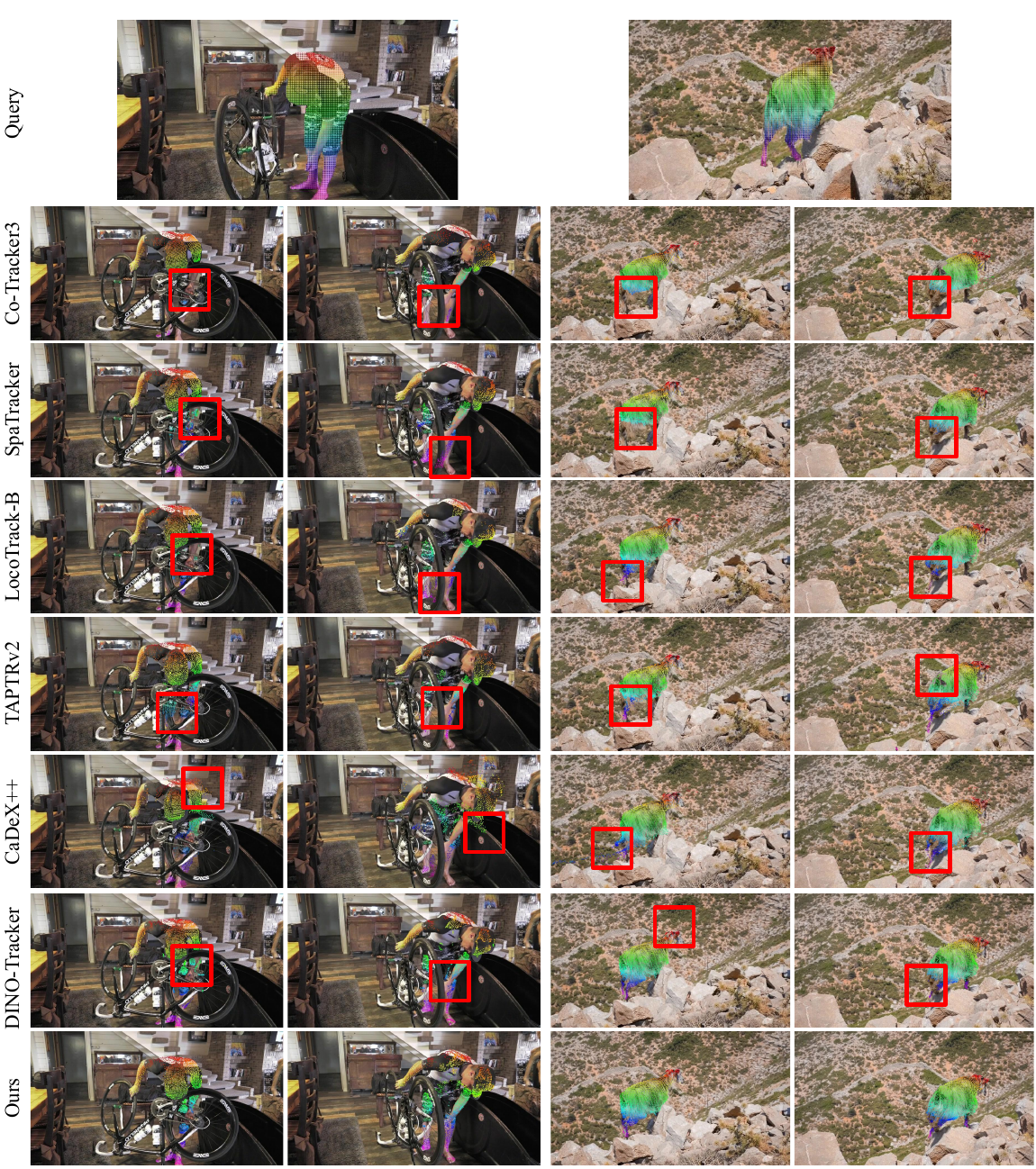}
    \caption{We evaluate our method against state-of-the-art approaches, including feed-forward models (Co-Tracker3~\cite{karaev24cotracker3}, SpatialTracker~\cite{xiao2024spatialtracker}, LocoTrack-B~\cite{cho2024local}, TAPTRv2~\cite{taptr}) and test-time training methods (CaDex++\cite{song2024track}, DINO-Tracker\cite{dino_tracker_2024}). Experiments are conducted on the \textit{bike-packing} and \textit{goat} scenes from TAPVid-DAVIS, with tracking at 256×256 resolution and visualizations at the original resolution. More qualitative results are provided in the supplementary material.}
    \label{fig:qualitative}
\end{figure*}

\subsection{Joint Flow and Long-term Correspondence Integration}
\label{sec:correspondence_tracking}
While flow integration can partially mitigate drift and produce smooth trajectories, accumulated errors still lead to drift over longer time spans. Moreover, in cases where an object disappears and reappears after some time, the optical flow method may struggle to track the point. To tackle these issues, we propose to integrate long-term correspondence into our flow-based prediction framework. 



We train a feature extractor $\mathbf{\Phi}_{\Delta}$ and heatmap refiner $\bm{\mathcal{R}}$ of a long-term correspondence-based keypoint tracker based on DINO-Tracker~\cite{dino_tracker_2024} for the input video, with the optical flow as a self-supervised signal. For frame $i$, the feature map $\bm{\mathcal{F}}_i$ can be calculated as:
\begin{equation}
    \bm{\mathcal{F}}_i=\mathbf{\Phi}_{\text{DINO}}(I^i)+\mathbf{\Phi}_{\Delta}(I^i)
\end{equation}
After getting the query feature $\bm{f}_{\text{query}}$ by sampling on the query point in $\bm{p}_0$, long-term predictions $\bm{p}_i$ are generated by applying \text{SoftArgMax} on the refined heatmap:
\begin{equation}
    \bm{p}_{i}=\text{SoftArgMax}(\bm{\mathcal{R}}(\bm{f}_{\text{query}}\cdot\bm{\mathcal{F}}_{i}))
\end{equation}
To avoid the negative impact of incorrect correspondences, we only require high-confidence keypoints. 
Thus, we only select those points with a cosine similarity greater than a threshold as keypoints. 
We incorporate these points into our probabilistic integration framework, as described by the following equation:
\begin{equation}
\bm{\mu}_i^{\text{key}} = 
\bm{p}_i,  \text{if } \bm{\mathcal{F}}_i(\bm{p}_i) \cdot \bm{f}_{\text{query}} > \rho, \sigma_i^{\text{key}}=1\\
\end{equation}
where $\rho = 0.7$, same as DINO-Tracker~\cite{dino_tracker_2024}.
Specifically, whenever valid keypoints from the long-term correspondence are available, we treat them as another source of noisy observations besides optical flow. In this way, we can jointly integrate the flow prediction and long-term keypoint in our probabilistic integration framework (Sec.~\ref{sec:kalman_filtering}) to yield a final optimal estimation of the point's location. Formally, let \(\bm{\mu}_i\) and \(\sigma_i\) represent the mean and variance from flow integration, and \(\bm{\mu}_i^{\text{key}}\) and \(\sigma_i^{\text{key}}\) represent the mean and variance from the key point observations. Then the combined estimates \(\bm{\mu}_i^{\text{final}}\) and \(\sigma_i^{\text{final}}\) are computed as:

\begin{equation}
\bm{\mu}_i^{\text{final}} = \frac{\bm{\mu}_i / \sigma_i^2 + \bm{\mu}_i^{\text{key}} / (\sigma_i^{\text{key}})^2}{1 / \sigma_i^2 + 1 / (\sigma_i^{\text{key}})^2},
\end{equation}
\begin{equation}
\sigma_i^{\text{final}} = \frac{1}{1 / \sigma_i^2 + 1 / (\sigma_i^{\text{key}})^2}.
\end{equation}

The final prediction for the point location is given by $\hat{\bm{p}}^t=\bm{\mu}_i^{\text{final}}$,  which indicates that our method's result has maximum likelihood within the final prediction distribution. If neither the optical flow nor the long-term key point is valid in the current frame, the point is marked as \textbf{occluded}. By incorporating long-term key points, our approach mitigates drift, effectively aligning flow estimation with long-term keypoints to maintain trajectory accuracy. Moreover, it enables the model to re-localize points that have temporarily disappeared and reappeared in different locations and can track through sudden scene transitions. As a result, the flow-based predictions continue to refine the point’s trajectory, ensuring accurate and smooth tracking across frames, as demonstrated in Fig.~\ref{fig:empty_flow}.

\section{Experiments}
\subsection{Experiment Setup}
\textbf{Dataset:} We evaluate our method on the following datasets from TAP-Vid~\cite{doersch2022tap} and BADJA~\cite{biggs2018creatures}:
\begin{itemize}
    \item \textbf{TAPVid-DAVIS}, a real-world dataset comprising 30 256px videos from DAVIS 2017~\cite{Pont-Tuset_arXiv_2017}. Each video contains between 34 and 104 RGB frames, capturing both camera movements and dynamic scene motions. We employ both the \textbf{query-first} mode, where all query points come from the first frame, and the \textbf{query-strided} mode, where a query is performed every 5 frames, for evaluation on DAVIS.
    \item \textbf{TAPVid-Kinetics}, includes 1,189 videos, each with 250 frames at 256px resolution from  Kinetics-700-2020~\cite{carreira2017quo}. The dataset predominantly focuses on human activity, with both camera and object motion. Since our method includes test-time optimization steps, we use the subset of 100 videos sampled by Omnimotion~\cite{wang2023tracking} and the query-first mode for evaluation.\
    \item \textbf{BADJA}, consists of nine videos, at 480px resolution, showcasing animal movements in nature, with ground truth information for keypoint locations.
\end{itemize}

\vspace{3pt}
\noindent\textbf{Metrics:} In accordance with the TAP-Vid~\cite{doersch2022tap} benchmark, we use the following metrics:
\begin{itemize}
    \item $\delta_{avg}^x$ measures the percentage of visible points that are tracked within a specific pixel error from the ground truth, which is evaluated over five thresholds: 
    \{1,2,4,8,16\} pixels, with the final score being the average fraction of points within these distances.
    \item \textbf{Occlusion Accuracy (OA)} measures the fraction of points with correct visibility predictions in each frame, including both visible and occluded points.
    \item \textbf{Average Jaccard (AJ)} measures both position and occlusion accuracy based on \(\delta^x_{avg}\) thresholds, which assesses the ratio of correctly predicted visible points to false predicted points.
\end{itemize}
For evaluating BADJA~\cite{biggs2018creatures},the following metrics are used:
\begin{itemize}
    \item $\delta^{seg}$ measures the percentage of predicted points that lie within the distance of $0.2\sqrt{A}$ from the ground-truth position, where $A$ is the area of the object mask.
    \item $\delta^{3px}$ measure the accuracy within a threshold of 3px.
   
\end{itemize}

\subsection{Comparisons}
\textbf{Baselines} We compare our \name to state-of-the-art methods:
\textbf{1)} \textit{optimization-based} trackers such as Omnimotion~\cite{wang2023tracking}, CaDeX++~\cite{song2024track}, DecoMotion~\cite{li2024decomposition} and DINO-Tracker~\cite{dino_tracker_2024}. Note that optimization-based methods do not require model training on labeled datasets. 
\textbf{2)} \textit{feedforward trackers trained in a supervised manner}, including PIPs~\cite{harley2022particle}, TAP-Net~\cite{doersch2022tap}, TAPIR~\cite{doersch2023tapir}, CoTracker~\cite{karaev2023cotracker},  SpatialTracker~\cite{xiao2024spatialtracker}, BootsTAP~\cite{doersch2024bootstap}, CoTracker3~\cite{karaev24cotracker3}, LocoTrack-B~\cite{cho2024local} and TAPTRv2~\cite{li2024taptrv2}. We additionally incorporate MFT~\cite{neoral2024mft}, which leverages RAFT to directly obtain trajectory predictions. 

\vspace{3pt}
\noindent\textbf{Quantitative comparisons}
Tab.~\ref{tab:performance_unsupervised} and Tab.~\ref{tab:performance} compare our method against state-of-the-art trackers on the TAP-Vid and BADJA benchmarks, where our approach achieves the highest \(\delta_{avg}^x\) across all datasets, demonstrating superior precision in tracking visible points. We attribute this to our probabilistic model, which enhances tracking accuracy by integrating optical flow with long-term correspondence, resulting in smoother and more precise trajectory segments. Additionally, in terms of Occlusion Accuracy (OA) and Average Jaccard (AJ), our method performs on par with the best approaches, effectively handling occlusions while maintaining geometric consistency. Notably, on BADJA, our method outperforms all other trackers.

Besides, among tracking methods that require test-time training, our method achieves the best performance across all metrics, demonstrating its robustness in both position and occlusion tracking. Additionally, our approach does not predict occlusion via trajectory agreement, making it a more efficient approach than DINO-Tracker~\cite{dino_tracker_2024}; refer to Appendix~\ref{sec:speed} for more discussion.



\vspace{3pt}
\noindent\textbf{Qualitative results} 

In Fig.~\ref{fig:qualitative}, our method could consistently trace the target points, even in challenging scenarios where objects frequently disappear and reappear. For instance, in the \textit{bike-packing} sequence, our tracking points are placed on the person. While some methods (CoTracker3, SpatialTracker, TAPTRv2, CaDeX++) occasionally mis-track these points onto the bike or background, or fail to capture finer details such as the hands (DINO-Tracker, LocoTrack), our approach maintains precise tracking throughout the video. In the \textit{goat} sequence, where the hooves frequently cross and obscure each other, our method reliably tracks the target points, remaining unaffected by overlapping limbs. We refer readers to our supplementary materials for more visual results.

Our method successfully locates the positions of most visible points at each time step through seamless integration of key components. The hybrid filter prevents drift during occlusions (e.g., the bike in bike-packing, overlapping legs in goat), while long-term keypoints enable reliable re-identification. Our probabilistic framework anchors these keypoints, using temporal-aware optical flow to track less distinctive points, ensuring each point is located to the greatest extent possible.



\subsection{Ablation study}
Next, we show an ablation study on different components of our framework using TapVid-DAVIS. Specifically, \textit{w/o keypoint} removes joint integration with long-term key points, using only flow integration results. \textit{w/o geo-aware} omits filtering by the geometry-aware feature. \textit{w/o mask} applies rough flow predictions without object-level filtering. \textit{w/o probabilistic} replaces probabilistic integration with selecting the prediction of the lowest $\sigma$ as the final result.

\begin{figure}[t]
    \centering
    
    \includegraphics[width=0.5\textwidth]{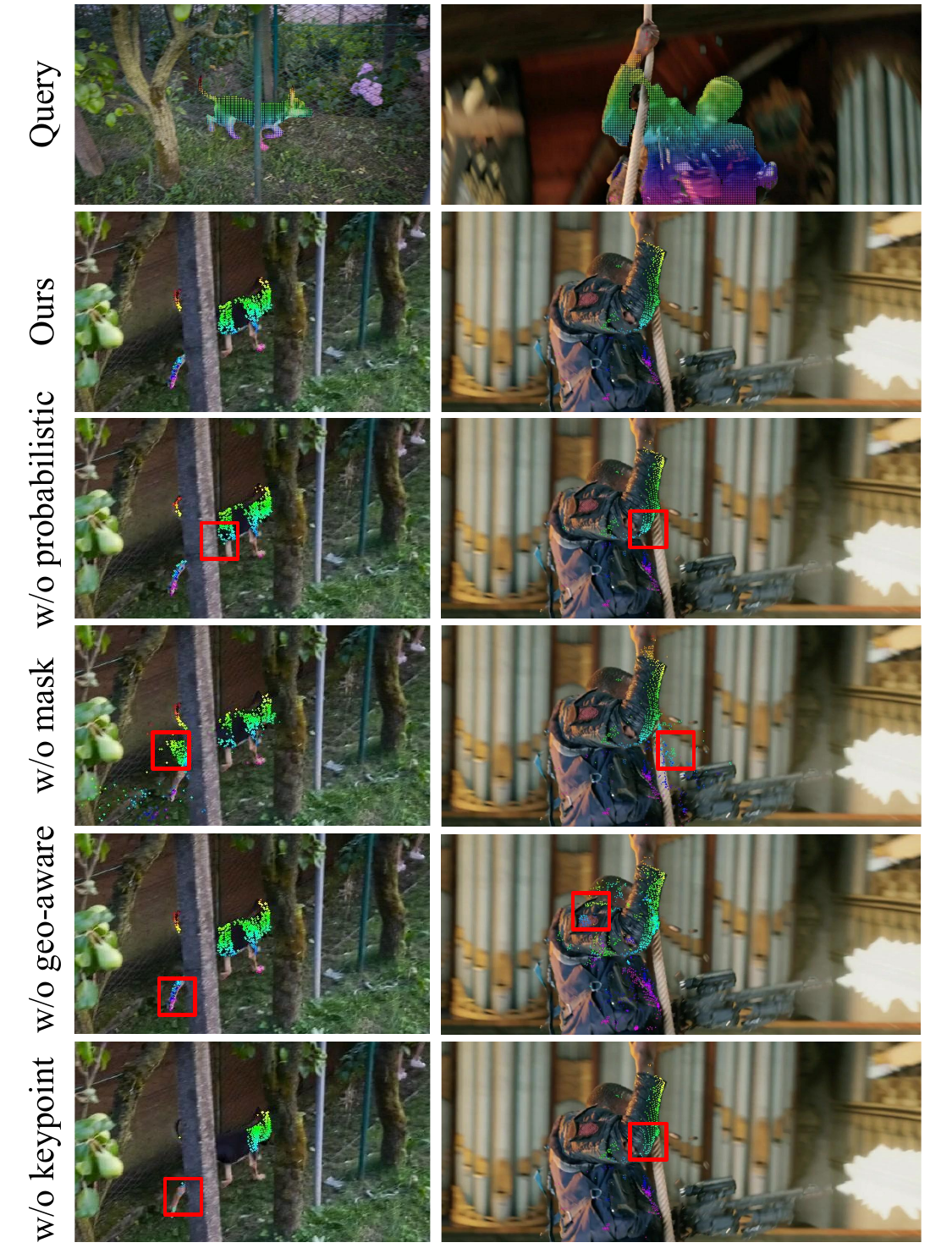}
    \vspace{-20pt}
    \caption{Ablation study on different components. 
    }
    \vspace{-20pt}
    \label{fig:ablation}
\end{figure}

As shown in Tab.\ref{tab:ablation} and Fig.\ref{fig:ablation}, mask filtering effectively removes incorrect flow predictions, significantly improving precision and occlusion handling. Long-term key points enable trajectory recovery beyond optical flow, leading to substantial performance gains. Probabilistic integration further enhances positional accuracy by reducing uncertainty. Additionally, the geometry-aware feature mitigates misalignment in visually similar regions, improving occlusion handling accuracy.



\begin{table}[t]
    \centering
    \begin{small}
    \begin{tabular}{lcccccc}
        \toprule
        \multirow{2}{*}{Method} & \multicolumn{3}{c}{DAVIS-First} & \multicolumn{3}{c}{DAVIS-Strided} \\
        \cmidrule(lr){2-4} \cmidrule(lr){5-7}
         & $\delta_{avg}^x$ & OA & AJ & $\delta_{avg}^x$ & OA & AJ \\
        \cmidrule(lr){1-1}\cmidrule(lr){2-4}\cmidrule(lr){5-7}
        w/o key point & 71.2 & 79.2 & 47.8 & 74.6 & 87.6 & 62.8 \\
        w/o geo-aware & \textbf{77.6} & 85.7 & 60.0 & \textbf{80.8} & 88.4 & 63.3 \\
        w/o mask & 72.3 & 82.3 & 57.1 & 73.9 & 82.6 & 57.8 \\
        w/o probabilistic & 76.9 & 87.2 & 61.3 & 79.0 & 88.2 & 63.9 \\
        Ours Full& \textbf{77.6} & \textbf{87.3} & \textbf{62.0} & \textbf{80.8} & \textbf{88.7} & \textbf{65.3} \\
        \bottomrule
    \end{tabular}
    \end{small}
    \caption{Ablation on different components on TAP-Vid-DAVIS. The best results are \textbf{bolded}.}
    \label{tab:ablation}
\end{table}

\section{Conclusion and Future Work}
In this paper, we introduced a robust tracking framework that combines optical flow integration with long-term correspondence through probabilistic integration to achieve accurate and smooth point tracking in dynamic video sequences. By incorporating object-level filtering, bidirectional probabilistic integration, and geometry-aware feature extraction, our method effectively mitigates drift, handles occlusions, and re-localizes temporally disappearing points. Our method outperforms traditional methods in handling complex motions and extended time gaps, demonstrating the advantages of integrating short-term and long-term information for reliable tracking.

While our method provides robust tracking, its reliance on test-time training for keypoint extraction reduces efficiency compared to supervised approaches—a common limitation of optimization-based tracking methods. 
This dependency on test-time training arises due to the current feature extractor's insufficient resolution and lack of temporal awareness. Future improvements in high-resolution feature extraction could help avoid test-time training and improve differentiation between objects and regions, allowing for fully unsupervised and real-time dense tracking.


{
    \small
    \bibliographystyle{ieeenat_fullname}
    \bibliography{main}

\begin{thebibliography}{53}
\providecommand{\natexlab}[1]{#1}
\providecommand{\url}[1]{\texttt{#1}}
\expandafter\ifx\csname urlstyle\endcsname\relax
  \providecommand{\doi}[1]{doi: #1}\else
  \providecommand{\doi}{doi: \begingroup \urlstyle{rm}\Url}\fi

\bibitem[Aydemir et~al.(2024)Aydemir, Xie, and Güney]{aydemir2024visualfoundationmodelsachieve}
Görkay Aydemir, Weidi Xie, and Fatma Güney.
\newblock Can visual foundation models achieve long-term point tracking?, 2024.

\bibitem[Biggs et~al.(2018)Biggs, Roddick, Fitzgibbon, and Cipolla]{biggs2018creatures}
Benjamin Biggs, Thomas Roddick, Andrew Fitzgibbon, and Roberto Cipolla.
\newblock {C}reatures great and {SMAL}: {R}ecovering the shape and motion of animals from video.
\newblock In \emph{ACCV}, 2018.

\bibitem[Brox et~al.(2004)Brox, Bruhn, Papenberg, and Weickert]{brox2004high}
Thomas Brox, Andr{\'e}s Bruhn, Nils Papenberg, and Joachim Weickert.
\newblock High accuracy optical flow estimation based on a theory for warping.
\newblock In \emph{Computer Vision-ECCV 2004: 8th European Conference on Computer Vision, Prague, Czech Republic, May 11-14, 2004. Proceedings, Part IV 8}, pages 25--36. Springer, 2004.

\bibitem[Brox et~al.(2009)Brox, Bregler, and Malik]{brox2009large}
Thomas Brox, Christoph Bregler, and Jitendra Malik.
\newblock Large displacement optical flow.
\newblock In \emph{2009 IEEE Conference on Computer Vision and Pattern Recognition}, pages 41--48. IEEE, 2009.

\bibitem[Carion et~al.(2020)Carion, Massa, Synnaeve, Usunier, Kirillov, and Zagoruyko]{carion2020endtoendobjectdetectiontransformers}
Nicolas Carion, Francisco Massa, Gabriel Synnaeve, Nicolas Usunier, Alexander Kirillov, and Sergey Zagoruyko.
\newblock End-to-end object detection with transformers, 2020.

\bibitem[Caron et~al.(2021)Caron, Touvron, Misra, J{\'e}gou, Mairal, Bojanowski, and Joulin]{caron2021emerging}
Mathilde Caron, Hugo Touvron, Ishan Misra, Herv{\'e} J{\'e}gou, Julien Mairal, Piotr Bojanowski, and Armand Joulin.
\newblock Emerging properties in self-supervised vision transformers.
\newblock In \emph{Proceedings of the IEEE/CVF international conference on computer vision}, pages 9650--9660, 2021.

\bibitem[Carreira and Zisserman(2017)]{carreira2017quo}
Joao Carreira and Andrew Zisserman.
\newblock Quo vadis, action recognition? a new model and the kinetics dataset.
\newblock In \emph{proceedings of the IEEE Conference on Computer Vision and Pattern Recognition}, pages 6299--6308, 2017.

\bibitem[Cheng et~al.(2024)Cheng, Deng, Harley, Zhu, and Guibas]{cheng2024zero}
Xinle Cheng, Congyue Deng, Adam Harley, Yixin Zhu, and Leonidas Guibas.
\newblock Zero-shot image feature consensus with deep functional maps.
\newblock \emph{arXiv preprint arXiv:2403.12038}, 2024.

\bibitem[Cho et~al.(2021)Cho, Hong, Jeon, Lee, Sohn, and Kim]{cho2021cats}
Seokju Cho, Sunghwan Hong, Sangryul Jeon, Yunsung Lee, Kwanghoon Sohn, and Seungryong Kim.
\newblock Cats: Cost aggregation transformers for visual correspondence.
\newblock \emph{Advances in Neural Information Processing Systems}, 34:\penalty0 9011--9023, 2021.

\bibitem[Cho et~al.(2024{\natexlab{a}})Cho, Huang, Kim, and Lee]{cho2024flowtrack}
Seokju Cho, Jiahui Huang, Seungryong Kim, and Joon-Young Lee.
\newblock Flowtrack: Revisiting optical flow for long-range dense tracking.
\newblock In \emph{Proceedings of the IEEE/CVF Conference on Computer Vision and Pattern Recognition}, pages 19268--19277, 2024{\natexlab{a}}.

\bibitem[Cho et~al.(2024{\natexlab{b}})Cho, Huang, Nam, An, Kim, and Lee]{cho2024local}
Seokju Cho, Jiahui Huang, Jisu Nam, Honggyu An, Seungryong Kim, and Joon-Young Lee.
\newblock Local all-pair correspondence for point tracking.
\newblock \emph{arXiv preprint arXiv:2407.15420}, 2024{\natexlab{b}}.

\bibitem[Doersch et~al.(2022)Doersch, Gupta, Markeeva, Recasens, Smaira, Aytar, Carreira, Zisserman, and Yang]{doersch2022tap}
Carl Doersch, Ankush Gupta, Larisa Markeeva, Adria Recasens, Lucas Smaira, Yusuf Aytar, Joao Carreira, Andrew Zisserman, and Yi Yang.
\newblock {TAP}-vid: A benchmark for tracking any point in a video.
\newblock \emph{Advances in Neural Information Processing Systems}, 35:\penalty0 13610--13626, 2022.

\bibitem[Doersch et~al.(2023)Doersch, Yang, Vecerik, Gokay, Gupta, Aytar, Carreira, and Zisserman]{doersch2023tapir}
Carl Doersch, Yi Yang, Mel Vecerik, Dilara Gokay, Ankush Gupta, Yusuf Aytar, Joao Carreira, and Andrew Zisserman.
\newblock Tapir: Tracking any point with per-frame initialization and temporal refinement.
\newblock In \emph{Proceedings of the IEEE/CVF International Conference on Computer Vision}, pages 10061--10072, 2023.

\bibitem[Doersch et~al.(2024)Doersch, Luc, Yang, Gokay, Koppula, Gupta, Heyward, Rocco, Goroshin, Carreira, and Zisserman]{doersch2024bootstap}
Carl Doersch, Pauline Luc, Yi Yang, Dilara Gokay, Skanda Koppula, Ankush Gupta, Joseph Heyward, Ignacio Rocco, Ross Goroshin, Jo{\~a}o Carreira, and Andrew Zisserman.
\newblock {BootsTAP}: Bootstrapped training for tracking-any-point.
\newblock \emph{Asian Conference on Computer Vision}, 2024.

\bibitem[Dosovitskiy et~al.(2015)Dosovitskiy, Fischer, Ilg, Hausser, Hazirbas, Golkov, Van Der~Smagt, Cremers, and Brox]{dosovitskiy2015flownet}
Alexey Dosovitskiy, Philipp Fischer, Eddy Ilg, Philip Hausser, Caner Hazirbas, Vladimir Golkov, Patrick Van Der~Smagt, Daniel Cremers, and Thomas Brox.
\newblock Flownet: Learning optical flow with convolutional networks.
\newblock In \emph{Proceedings of the IEEE international conference on computer vision}, pages 2758--2766, 2015.

\bibitem[Gu et~al.(2024)Gu, Zhou, Wu, Yu, Liu, Zhao, Wu, Zhang, Shou, and Tang]{Gu_2024_CVPR}
Yuchao Gu, Yipin Zhou, Bichen Wu, Licheng Yu, Jia-Wei Liu, Rui Zhao, Jay~Zhangjie Wu, David~Junhao Zhang, Mike~Zheng Shou, and Kevin Tang.
\newblock Videoswap: Customized video subject swapping with interactive semantic point correspondence.
\newblock In \emph{Proceedings of the IEEE/CVF Conference on Computer Vision and Pattern Recognition (CVPR)}, pages 7621--7630, 2024.

\bibitem[Harley et~al.(2022)Harley, Fang, and Fragkiadaki]{harley2022particle}
Adam~W Harley, Zhaoyuan Fang, and Katerina Fragkiadaki.
\newblock Particle video revisited: Tracking through occlusions using point trajectories.
\newblock In \emph{European Conference on Computer Vision}, pages 59--75. Springer, 2022.

\bibitem[Hedlin et~al.(2024)Hedlin, Sharma, Mahajan, Isack, Kar, Tagliasacchi, and Yi]{hedlin2024unsupervised}
Eric Hedlin, Gopal Sharma, Shweta Mahajan, Hossam Isack, Abhishek Kar, Andrea Tagliasacchi, and Kwang~Moo Yi.
\newblock Unsupervised semantic correspondence using stable diffusion.
\newblock \emph{Advances in Neural Information Processing Systems}, 36, 2024.

\bibitem[Horn and Schunck(1981)]{horn1981determining}
Berthold~KP Horn and Brian~G Schunck.
\newblock Determining optical flow.
\newblock \emph{Artificial intelligence}, 17\penalty0 (1-3):\penalty0 185--203, 1981.

\bibitem[Huang et~al.(2022)Huang, Shi, Zhang, Wang, Cheung, Qin, Dai, and Li]{huang2022flowformer}
Zhaoyang Huang, Xiaoyu Shi, Chao Zhang, Qiang Wang, Ka~Chun Cheung, Hongwei Qin, Jifeng Dai, and Hongsheng Li.
\newblock Flowformer: A transformer architecture for optical flow.
\newblock In \emph{European conference on computer vision}, pages 668--685. Springer, 2022.

\bibitem[Ilg et~al.(2017)Ilg, Mayer, Saikia, Keuper, Dosovitskiy, and Brox]{ilg2017flownet}
Eddy Ilg, Nikolaus Mayer, Tonmoy Saikia, Margret Keuper, Alexey Dosovitskiy, and Thomas Brox.
\newblock Flownet 2.0: Evolution of optical flow estimation with deep networks.
\newblock In \emph{Proceedings of the IEEE conference on computer vision and pattern recognition}, pages 2462--2470, 2017.

\bibitem[Kalman(1960)]{kalman1960new}
Rudolph~Emil Kalman.
\newblock A new approach to linear filtering and prediction problems.
\newblock 1960.

\bibitem[Karaev et~al.(2023)Karaev, Rocco, Graham, Neverova, Vedaldi, and Rupprecht]{karaev2023cotracker}
Nikita Karaev, Ignacio Rocco, Benjamin Graham, Natalia Neverova, Andrea Vedaldi, and Christian Rupprecht.
\newblock Cotracker: It is better to track together.
\newblock \emph{arXiv preprint arXiv:2307.07635}, 2023.

\bibitem[Karaev et~al.(2024)Karaev, Makarov, Wang, Neverova, Vedaldi, and Rupprecht]{karaev24cotracker3}
Nikita Karaev, Iurii Makarov, Jianyuan Wang, Natalia Neverova, Andrea Vedaldi, and Christian Rupprecht.
\newblock Cotracker3: Simpler and better point tracking by pseudo-labelling real videos.
\newblock In \emph{Proc. {arXiv:2410.11831}}, 2024.

\bibitem[Lei et~al.(2024)Lei, Weng, Harley, Guibas, and Daniilidis]{lei2024moscadynamicgaussianfusion}
Jiahui Lei, Yijia Weng, Adam Harley, Leonidas Guibas, and Kostas Daniilidis.
\newblock Mosca: Dynamic gaussian fusion from casual videos via 4d motion scaffolds, 2024.

\bibitem[Li et~al.(2024{\natexlab{a}})Li, Zhang, Liu, Zeng, Li, Ren, Li, and Zhang]{li2024taptrv2}
Hongyang Li, Hao Zhang, Shilong Liu, Zhaoyang Zeng, Feng Li, Tianhe Ren, Bohan Li, and Lei Zhang.
\newblock Taptrv2: Attention-based position update improves tracking any point.
\newblock \emph{arXiv preprint arXiv:2407.16291}, 2024{\natexlab{a}}.

\bibitem[Li et~al.(2024{\natexlab{b}})Li, Zhang, Liu, Zeng, Ren, Li, and Zhang]{taptr}
Hongyang Li, Hao Zhang, Shilong Liu, Zhaoyang Zeng, Tianhe Ren, Feng Li, and Lei Zhang.
\newblock Taptr: Tracking any point with transformers as detection.
\newblock In \emph{Proceedings of the IEEE/CVF European Conference on Computer Vision}, 2024{\natexlab{b}}.

\bibitem[Li and Liu(2024)]{li2024decomposition}
Rui Li and Dong Liu.
\newblock Decomposition betters tracking everything everywhere.
\newblock \emph{arXiv preprint arXiv:2407.06531}, 2024.

\bibitem[Lowe(2004)]{lowe2004sift}
David~G. Lowe.
\newblock Distinctive image features from scale-invariant keypoints.
\newblock \emph{International Journal of Computer Vision}, 60:\penalty0 91--110, 2004.

\bibitem[Lucas and Kanade(1981)]{lucas1981iterative}
Bruce~D Lucas and Takeo Kanade.
\newblock An iterative image registration technique with an application to stereo vision.
\newblock In \emph{IJCAI'81: 7th international joint conference on Artificial intelligence}, pages 674--679, 1981.

\bibitem[Luo et~al.(2024)Luo, Dunlap, Park, Holynski, and Darrell]{luo2024diffusion}
Grace Luo, Lisa Dunlap, Dong~Huk Park, Aleksander Holynski, and Trevor Darrell.
\newblock Diffusion hyperfeatures: Searching through time and space for semantic correspondence.
\newblock \emph{Advances in Neural Information Processing Systems}, 36, 2024.

\bibitem[Melekhov et~al.(2019)Melekhov, Tiulpin, Sattler, Pollefeys, Rahtu, and Kannala]{melekhov2019dgc}
Iaroslav Melekhov, Aleksei Tiulpin, Torsten Sattler, Marc Pollefeys, Esa Rahtu, and Juho Kannala.
\newblock Dgc-net: Dense geometric correspondence network.
\newblock In \emph{2019 IEEE Winter Conference on Applications of Computer Vision (WACV)}, pages 1034--1042. IEEE, 2019.

\bibitem[Neoral et~al.(2024)Neoral, {\v{S}}er{\`y}ch, and Matas]{neoral2024mft}
Michal Neoral, Jon{\'a}{\v{s}} {\v{S}}er{\`y}ch, and Ji{\v{r}}{\'\i} Matas.
\newblock Mft: Long-term tracking of every pixel.
\newblock In \emph{Proceedings of the IEEE/CVF Winter Conference on Applications of Computer Vision}, pages 6837--6847, 2024.

\bibitem[Oquab et~al.(2023)Oquab, Darcet, Moutakanni, Vo, Szafraniec, Khalidov, Fernandez, Haziza, Massa, El-Nouby, et~al.]{oquab2023dinov2}
Maxime Oquab, Timoth{\'e}e Darcet, Th{\'e}o Moutakanni, Huy Vo, Marc Szafraniec, Vasil Khalidov, Pierre Fernandez, Daniel Haziza, Francisco Massa, Alaaeldin El-Nouby, et~al.
\newblock Dinov2: Learning robust visual features without supervision.
\newblock \emph{arXiv preprint arXiv:2304.07193}, 2023.

\bibitem[Pont-Tuset et~al.(2017)Pont-Tuset, Perazzi, Caelles, Arbel\'aez, Sorkine-Hornung, and {Van Gool}]{Pont-Tuset_arXiv_2017}
Jordi Pont-Tuset, Federico Perazzi, Sergi Caelles, Pablo Arbel\'aez, Alexander Sorkine-Hornung, and Luc {Van Gool}.
\newblock The 2017 davis challenge on video object segmentation.
\newblock \emph{arXiv:1704.00675}, 2017.

\bibitem[Radford et~al.(2021)Radford, Kim, Hallacy, Ramesh, Goh, Agarwal, Sastry, Askell, Mishkin, Clark, et~al.]{radford2021learning}
Alec Radford, Jong~Wook Kim, Chris Hallacy, Aditya Ramesh, Gabriel Goh, Sandhini Agarwal, Girish Sastry, Amanda Askell, Pamela Mishkin, Jack Clark, et~al.
\newblock Learning transferable visual models from natural language supervision.
\newblock In \emph{International conference on machine learning}, pages 8748--8763. PMLR, 2021.

\bibitem[Ravi et~al.(2024)Ravi, Gabeur, Hu, Hu, Ryali, Ma, Khedr, R{\"a}dle, Rolland, Gustafson, Mintun, Pan, Alwala, Carion, Wu, Girshick, Doll{\'a}r, and Feichtenhofer]{ravi2024sam2}
Nikhila Ravi, Valentin Gabeur, Yuan-Ting Hu, Ronghang Hu, Chaitanya Ryali, Tengyu Ma, Haitham Khedr, Roman R{\"a}dle, Chloe Rolland, Laura Gustafson, Eric Mintun, Junting Pan, Kalyan~Vasudev Alwala, Nicolas Carion, Chao-Yuan Wu, Ross Girshick, Piotr Doll{\'a}r, and Christoph Feichtenhofer.
\newblock Sam 2: Segment anything in images and videos.
\newblock \emph{arXiv preprint arXiv:2408.00714}, 2024.

\bibitem[Rocco et~al.(2020)Rocco, Arandjelovi{\'c}, and Sivic]{rocco2020efficient}
Ignacio Rocco, Relja Arandjelovi{\'c}, and Josef Sivic.
\newblock Efficient neighbourhood consensus networks via submanifold sparse convolutions.
\newblock In \emph{Computer Vision--ECCV 2020: 16th European Conference, Glasgow, UK, August 23--28, 2020, Proceedings, Part IX 16}, pages 605--621. Springer, 2020.

\bibitem[Rombach et~al.(2022)Rombach, Blattmann, Lorenz, Esser, and Ommer]{rombach2022high}
Robin Rombach, Andreas Blattmann, Dominik Lorenz, Patrick Esser, and Bj{\"o}rn Ommer.
\newblock High-resolution image synthesis with latent diffusion models.
\newblock In \emph{Proceedings of the IEEE/CVF conference on computer vision and pattern recognition}, pages 10684--10695, 2022.

\bibitem[Rubinstein and Liu(2012)]{longrangemotion}
Michael Rubinstein and Ce Liu.
\newblock Towards longer long-range motion trajectories.
\newblock In \emph{Proceedings of the British Machine Vision Conference}, pages 53.1--53.11. BMVA Press, 2012.

\bibitem[Rublee et~al.(2011)Rublee, Rabaud, Konolige, and Bradski]{rublee2011orb}
Ethan Rublee, Vincent Rabaud, Kurt Konolige, and Gary Bradski.
\newblock Orb: An efficient alternative to sift or surf.
\newblock In \emph{2011 International conference on computer vision}, pages 2564--2571. Ieee, 2011.

\bibitem[Sand and Teller(2006)]{1641022}
P. Sand and S. Teller.
\newblock Particle video: Long-range motion estimation using point trajectories.
\newblock In \emph{2006 IEEE Computer Society Conference on Computer Vision and Pattern Recognition (CVPR'06)}, pages 2195--2202, 2006.

\bibitem[Song et~al.(2024)Song, Lei, Wang, Liu, and Daniilidis]{song2024track}
Yunzhou Song, Jiahui Lei, Ziyun Wang, Lingjie Liu, and Kostas Daniilidis.
\newblock Track everything everywhere fast and robustly.
\newblock \emph{arXiv preprint arXiv:2403.17931}, 2024.

\bibitem[Stearns et~al.(2024)Stearns, Harley, Uy, Dubost, Tombari, Wetzstein, and Guibas]{stearns2024dynamicgaussianmarblesnovel}
Colton Stearns, Adam Harley, Mikaela Uy, Florian Dubost, Federico Tombari, Gordon Wetzstein, and Leonidas Guibas.
\newblock Dynamic gaussian marbles for novel view synthesis of casual monocular videos, 2024.

\bibitem[Sun et~al.(2018)Sun, Yang, Liu, and Kautz]{sun2018pwc}
Deqing Sun, Xiaodong Yang, Ming-Yu Liu, and Jan Kautz.
\newblock Pwc-net: Cnns for optical flow using pyramid, warping, and cost volume.
\newblock In \emph{Proceedings of the IEEE conference on computer vision and pattern recognition}, pages 8934--8943, 2018.

\bibitem[Tang et~al.(2023)Tang, Jia, Wang, Phoo, and Hariharan]{tang2023emergent}
Luming Tang, Menglin Jia, Qianqian Wang, Cheng~Perng Phoo, and Bharath Hariharan.
\newblock Emergent correspondence from image diffusion.
\newblock \emph{Advances in Neural Information Processing Systems}, 36:\penalty0 1363--1389, 2023.

\bibitem[Teed and Deng(2020)]{teed2020raft}
Zachary Teed and Jia Deng.
\newblock Raft: Recurrent all-pairs field transforms for optical flow.
\newblock In \emph{Computer Vision--ECCV 2020: 16th European Conference, Glasgow, UK, August 23--28, 2020, Proceedings, Part II 16}, pages 402--419. Springer, 2020.

\bibitem[Truong et~al.(2020)Truong, Danelljan, and Timofte]{truong2020glu}
Prune Truong, Martin Danelljan, and Radu Timofte.
\newblock Glu-net: Global-local universal network for dense flow and correspondences.
\newblock In \emph{Proceedings of the IEEE/CVF conference on computer vision and pattern recognition}, pages 6258--6268, 2020.

\bibitem[Tumanyan et~al.(2024)Tumanyan, Singer, Bagon, and Dekel]{dino_tracker_2024}
Narek Tumanyan, Assaf Singer, Shai Bagon, and Tali Dekel.
\newblock Dino-tracker: Taming dino for self-supervised point tracking in a single video, 2024.

\bibitem[Wang et~al.(2023)Wang, Chang, Cai, Li, Hariharan, Holynski, and Snavely]{wang2023tracking}
Qianqian Wang, Yen-Yu Chang, Ruojin Cai, Zhengqi Li, Bharath Hariharan, Aleksander Holynski, and Noah Snavely.
\newblock Tracking everything everywhere all at once.
\newblock In \emph{Proceedings of the IEEE/CVF International Conference on Computer Vision}, pages 19795--19806, 2023.

\bibitem[Wang et~al.(2024)Wang, Ye, Gao, Austin, Li, and Kanazawa]{som2024}
Qianqian Wang, Vickie Ye, Hang Gao, Jake Austin, Zhengqi Li, and Angjoo Kanazawa.
\newblock Shape of motion: 4d reconstruction from a single video.
\newblock 2024.

\bibitem[Xiao et~al.(2024)Xiao, Wang, Zhang, Xue, Peng, Shen, and Zhou]{xiao2024spatialtracker}
Yuxi Xiao, Qianqian Wang, Shangzhan Zhang, Nan Xue, Sida Peng, Yujun Shen, and Xiaowei Zhou.
\newblock Spatialtracker: Tracking any 2d pixels in 3d space.
\newblock In \emph{Proceedings of the IEEE/CVF Conference on Computer Vision and Pattern Recognition}, pages 20406--20417, 2024.

\bibitem[Zhang et~al.(2024)Zhang, Herrmann, Hur, Chen, Jampani, Sun, and Yang]{zhang2024telling}
Junyi Zhang, Charles Herrmann, Junhwa Hur, Eric Chen, Varun Jampani, Deqing Sun, and Ming-Hsuan Yang.
\newblock Telling left from right: Identifying geometry-aware semantic correspondence.
\newblock In \emph{Proceedings of the IEEE/CVF Conference on Computer Vision and Pattern Recognition}, pages 3076--3085, 2024.

\end{thebibliography}
}
\clearpage
\setcounter{page}{1}
\maketitlesupplementary
\nocite{}

\setcounter{section}{0}
\setcounter{equation}{0}
\setcounter{figure}{0}
\section{Video Results}
Please refer to our Supplementary Webpage for the corresponding videos of images illustrated in the paper and more results on different data.
\section{Implementation Details}
\label{sec:detail}

\subsection{Hyperparameters}  
During the dual filtering stage, we apply different thresholds to predictions from flow and long-term keypoints.  
For long-term keypoints, we only need those with higher confidence to avoid mistakes. A prediction is first marked as invalid if the cosine similarity to the query point on the refined DINO feature falls below 0.7, following DINO-Tracker. It is then filtered following the same procedure by geometry-aware feature with the similarity threshold of 0.5, which is a common practice.
For predictions from flow, however, we want them to help track areas with less distinct features. Thus we use a threshold of 0.3 instead.  
These distinct thresholds allow flow to track featureless areas while ensuring that long-term keypoints do not drift into visually similar regions.
\textbf{Note that we use the same hyperparameters for all videos and our method don't require any hyperparameter tuning.}

\subsection{Flow Preparation} 
We utilize the RAFT~\cite{teed2020raft} model, as adopted by MFT~\cite{neoral2024mft}, as our flow estimation model.  
It takes two images, \(\bm{I}_j\) and \(\bm{I}_i\), captured at different times as input and outputs the flow map \(\bm{\mathcal{F}}_{ji}\),  
occlusion map \(\bm{\mathcal{O}}_{ji}\), and uncertainty map \(\bm{\mathcal{U}}_{ji}\).  
Since the uncertainty of an estimation can also be interpreted as its variance,  
the initial flow prediction from frame \(j\) to frame \(i\) at location \(\bm{p}\) is computed as follows:  
\begin{equation}
\begin{split}
    (\bm{f}_{ji},\mu_{ji}) =
    \begin{cases}(S(\bm{\mathcal{F}}_{ji}, \bm{p}),S(\bm{\mathcal{U}}_{ji},\bm{p})) &\text{if} (\bm{\mathcal{O}}_{ji},\bm{p})) > \rho\\
     None & \text{otherwise}
    \end{cases}
\end{split}
\end{equation}
where $\rho = 0.1$, and $S(Target, \bm{p})$ indicates sampling the target at location $\bm{p}$. The initial flow predictions are then set as input to flow integration. Following MFT, we adopt \(\{\infty, 1, 2, 4, 8, 16, 32\}\) as the time intervals, meaning that each frame's prediction is computed by combining results from these previous frames.

\subsection{Outlier Removal in Integration}  
Even after dual filtering, occasional erroneous predictions may persist.  
To ensure a more stable integration process and minimize the impact of these incorrect predictions, we discard rough predictions that deviate significantly from others. When a rough prediction is more than 10 pixels away from the most trustworthy one, we mark it as wrong prediction.
Denoting the input predictions as \(\{(\bm{f}_1, \bm{\mu}_1), (\bm{f}_2, \bm{\mu}_2), \dots, (\bm{f}_N, \bm{\mu}_N)\}\), we remove outliers using the following criterion:  
\begin{equation}
    (\bm{f}_{i},\bm{\mu}_{i}) =
    \begin{cases}
    (\bm{f}_{i},\bm{\mu}_{i}) &\text{if} \; \text{Norm}(\bm{f}_{i}-\bm{f}_{T}) < \rho_{dist}\\
    \text{None} &\text{otherwise}
    \end{cases}
\end{equation}
where $\rho_{dist}=10$, $T=argmin(\mu_i)$ and $\text{Norm}(\bm{x})$ measures the magnitude of a vector.

\section{Dense Inference}
As discussed in Sec.3.1 in the main paper, we utilize a geometry-aware feature extractor and a video mask generator for the dual-filter stage. While optical flow and geometry-aware features can be computed densely, generating masks for each pixel is both time-intensive and memory-intensive. To address this, we adopt an iterative approach to efficiently generate a set of masks that collectively cover all pixels, as described below:

\begin{algorithm}
\caption{Dense Mask Generation Algorithm}
\label{alg:dense_mask}

\begin{algorithmic}[1]
\STATE \textbf{Initialize:} Set all pixels as unassigned.
\WHILE{there exist unassigned pixels}
    \STATE Select the first unassigned pixel $p$.
    \STATE Generate a new mask $M$ starting from pixel $p$.
    \FOR{each pixel $q$ in $M$}
        \IF{$q$ is unassigned}
            \STATE Assign $q$ to mask $M$.
        \ENDIF
    \ENDFOR
\ENDWHILE
\STATE \textbf{Output:} All pixels assigned to corresponding masks.
\end{algorithmic}

\end{algorithm}

\noindent Subsequently, the dual filter can be applied to each pixel based on its corresponding mask.

\section{Training and Inference Speed}
\label{sec:speed}
Our methods is more than 20x faster than DINO-Tracker during the inference stage, while maintaining the same training time.

\noindent The total time consumed for our method includes the time for keypoint extraction, mask generation, geometry-aware feature extraction 
and probabilistic integration. During keypoint extraction, we follow DINO-Tracker~\cite{dino_tracker_2024} to train a delta-DINO model 
and a heatmap refiner, which takes about 1 hour for an 80-frame video on a single RTX 4090 GPU. 
We refer to DINO-Tracker~\cite{dino_tracker_2024} for more details. However, our method skips the time-consuming occlusion prediction and directly uses points 
with high cosine similarity as keypoints, which saves much time. 
The mask generation and geometry-aware feature extraction together takes about 2 minutes and the probabilistic integration takes about 1 minute for the same video.

\noindent In total, during the inference stage, the time spent tracking 3,000 points on a single object
in an 80-frame video is about 3 minutes, which is about \textbf{20x faster} than DINO-Tracker~\cite{dino_tracker_2024}. For dense inference, an additional 4 minutes may be required due to the increased number of masks generated, but our method remains more than 30x faster than DINO-Tracker~\cite{dino_tracker_2024}.

\noindent Although our method is comprised of several components, we provide convenient interface to run our method directly. Our code will be released upon publication.

\section{More Qualitative Results}
To further illustrate our methods' robustness. We conduct experiments on more challenging cases and show the qualitative results.

\begin{figure*}[h]
    \centering
    \includegraphics[width=\linewidth]{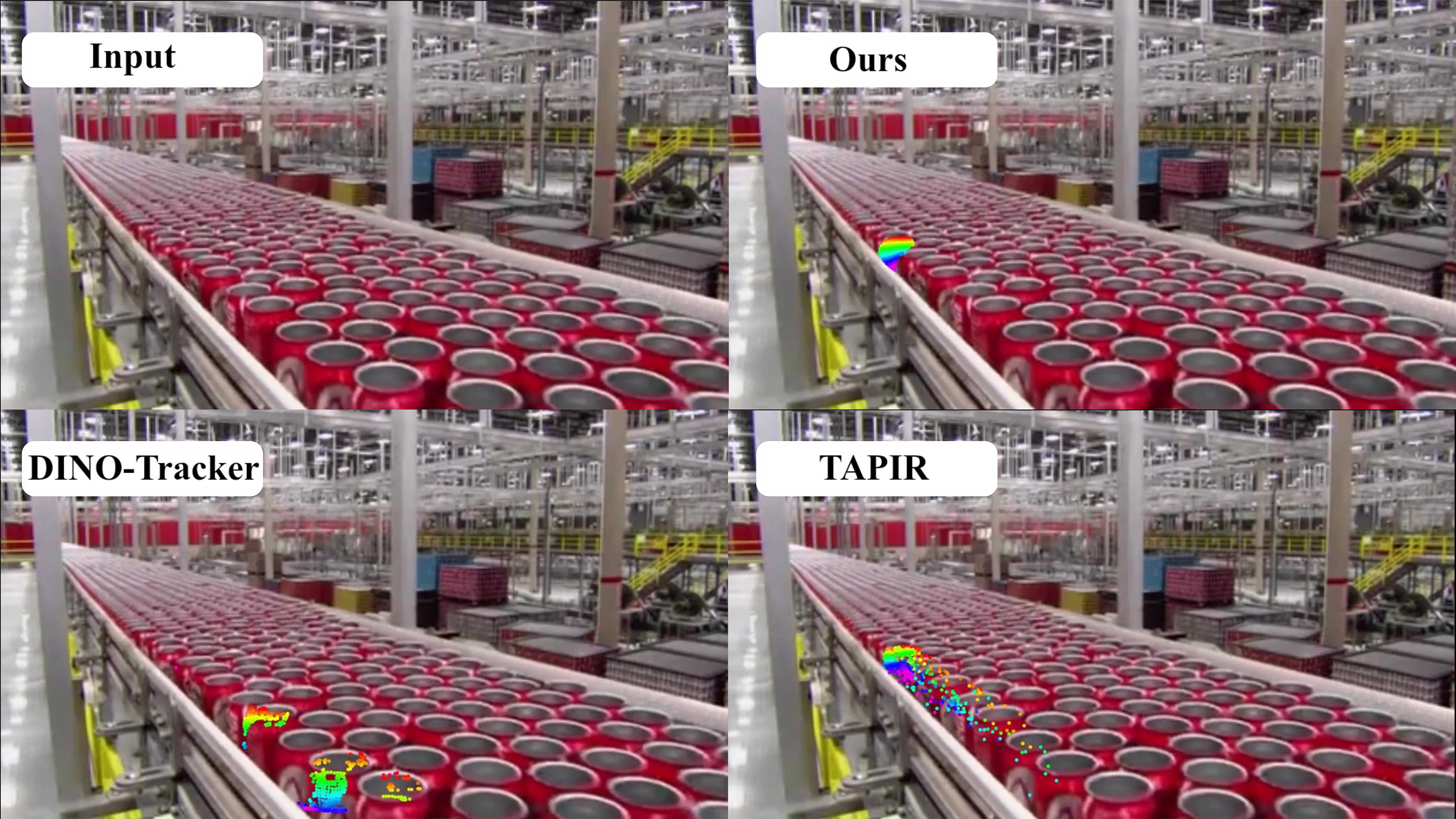} 
    \includegraphics[width=\linewidth]{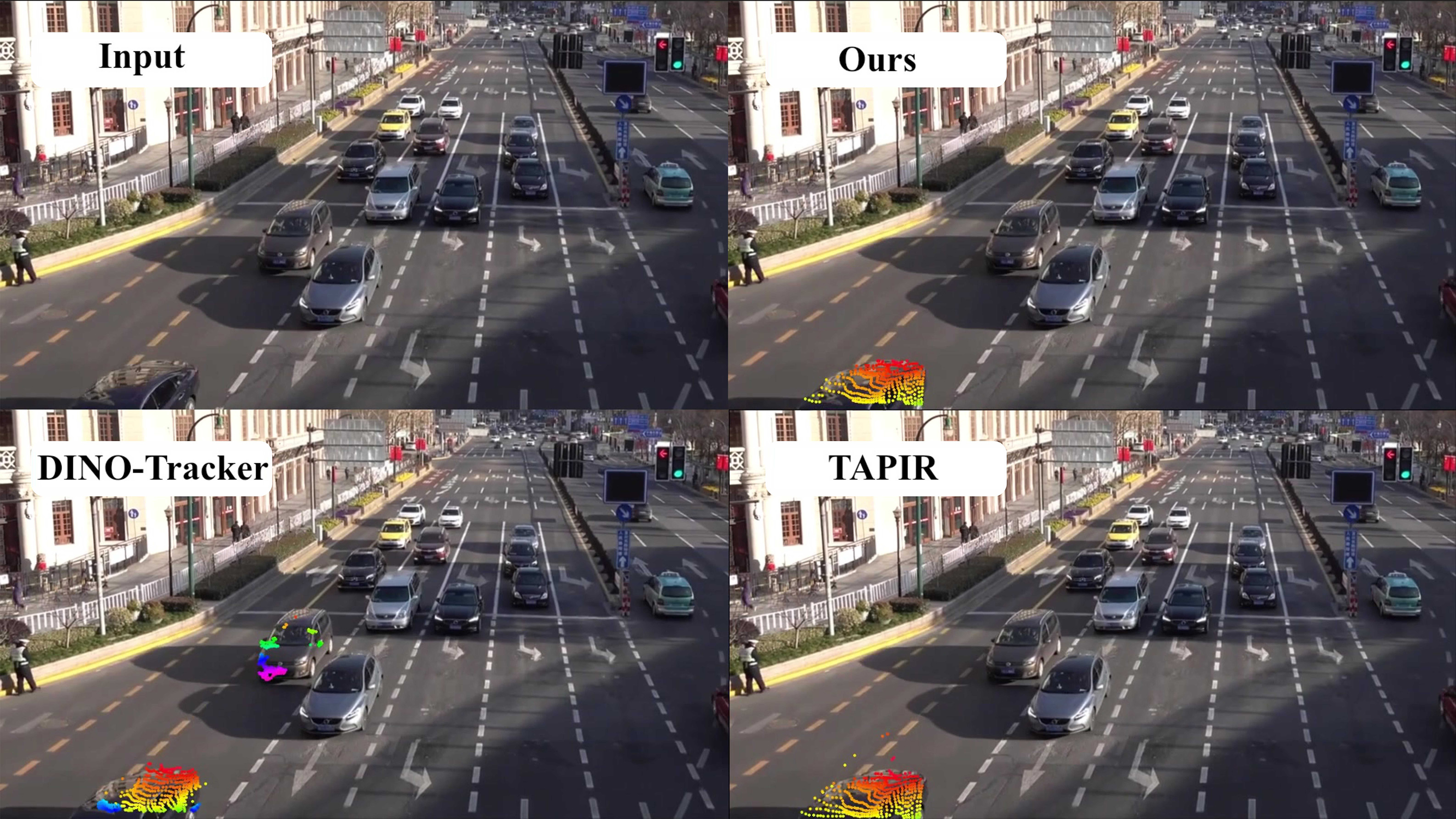} 
    \caption{Results of tracking a single object. While DINO-Tracker may mispredict parts onto similar objects and TAPIR can be disrupted by similar patterns, our method avoids these errors.} 
    \label{fig:case1}  
\end{figure*}

\begin{figure*}[h]
    \centering
    \includegraphics[width=\linewidth]{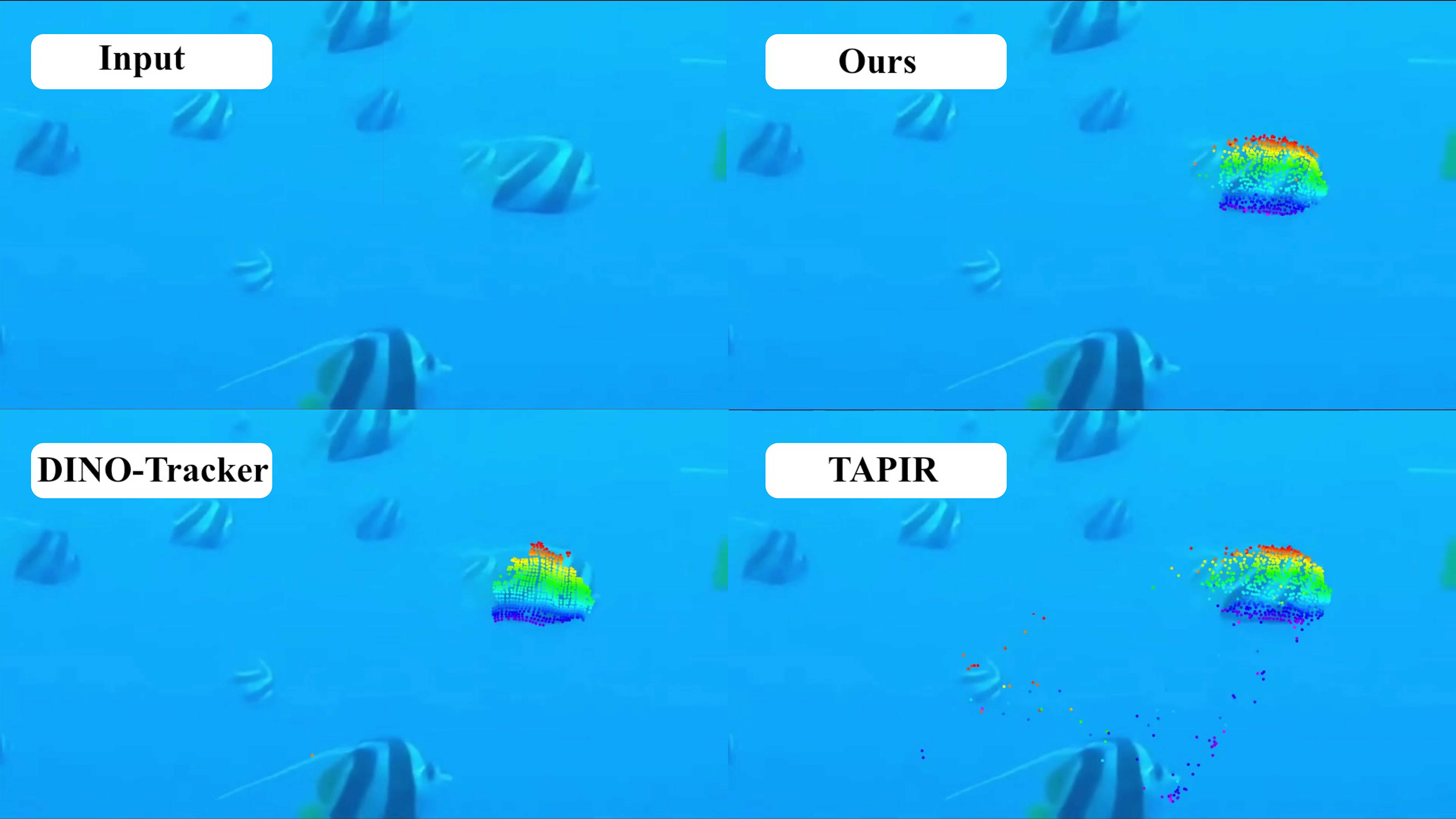} 
    \includegraphics[width=\linewidth]{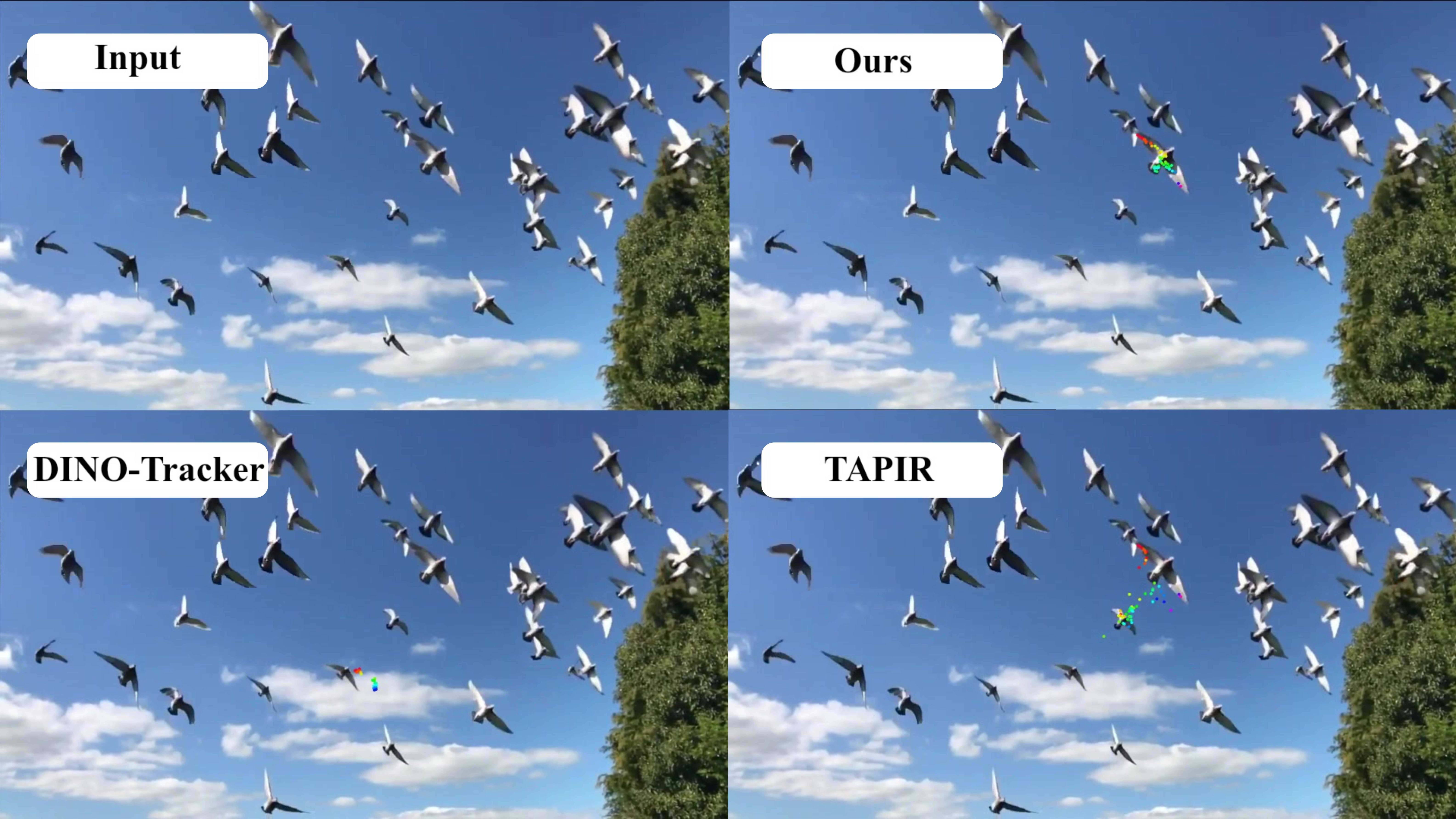} 
    \caption{Results of tracking a single object. While DINO-Tracker may lose some parts and TAPIR can be disrupted by multiple similar patterns, our method avoids these errors.} 
    \label{fig:case2}  
\end{figure*}

Some of the previous methods rely on computing a heatmap between the query point and the target frame. However, the per-frame heatmap lacks temporal-awareness and may confuse different objects. We address this issue by leveraging the mask and combining the heatmap with optical flow. As illustrated in Fig.~\ref{fig:case1} and Fig.~\ref{fig:case2}, by comparing the results of our method with DINO-Tracker~\cite{dino_tracker_2024} and TAPIR~\cite{doersch2023tapir}, we show that although our method also relies on per-frame heatmap to extract keypoints,our method has strong temporal-awareness and is able to tell between similar objects.


\begin{figure*}[h]
    \centering
    \includegraphics[width=\linewidth]{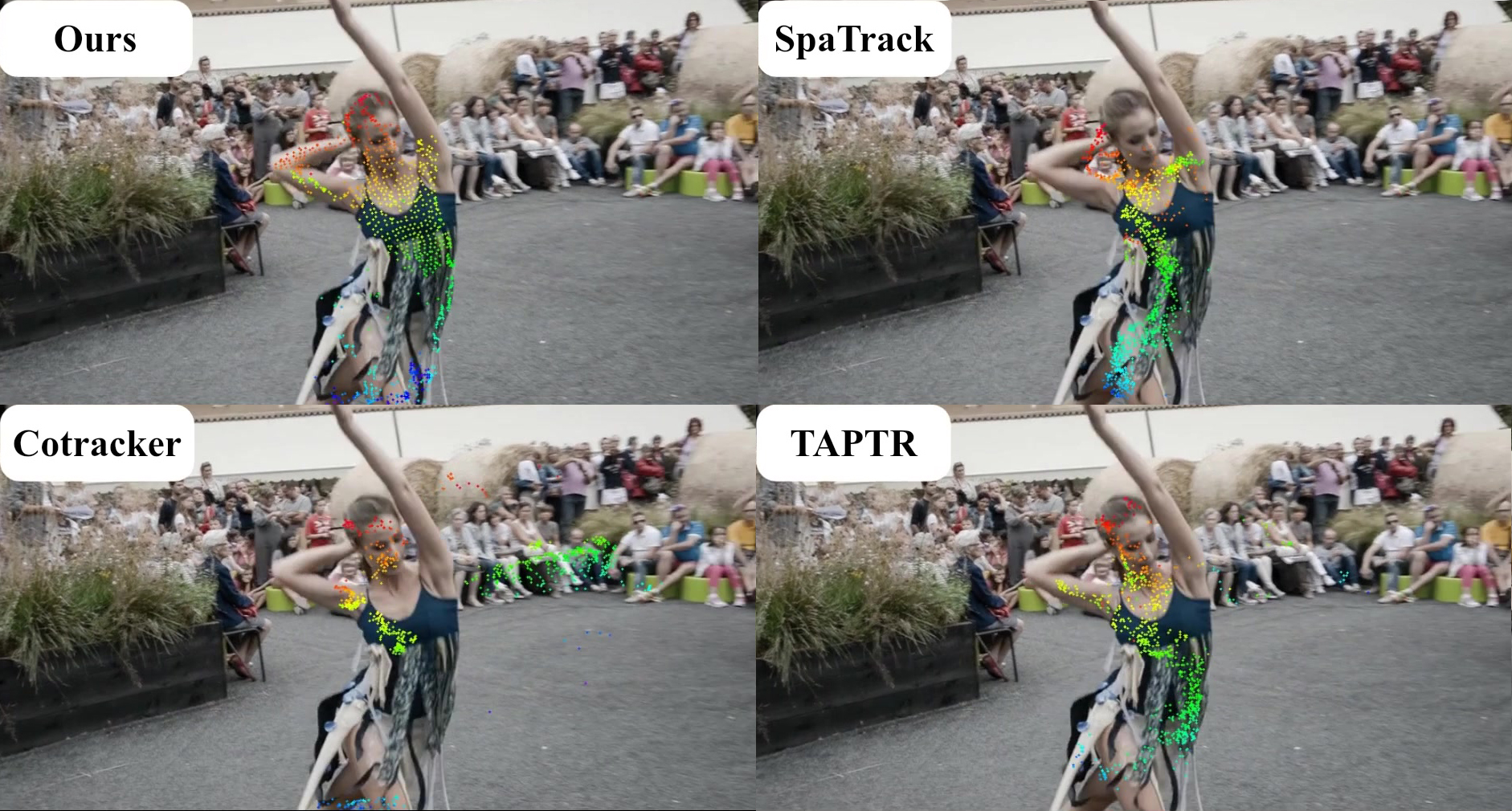} 
    \includegraphics[width=\linewidth]{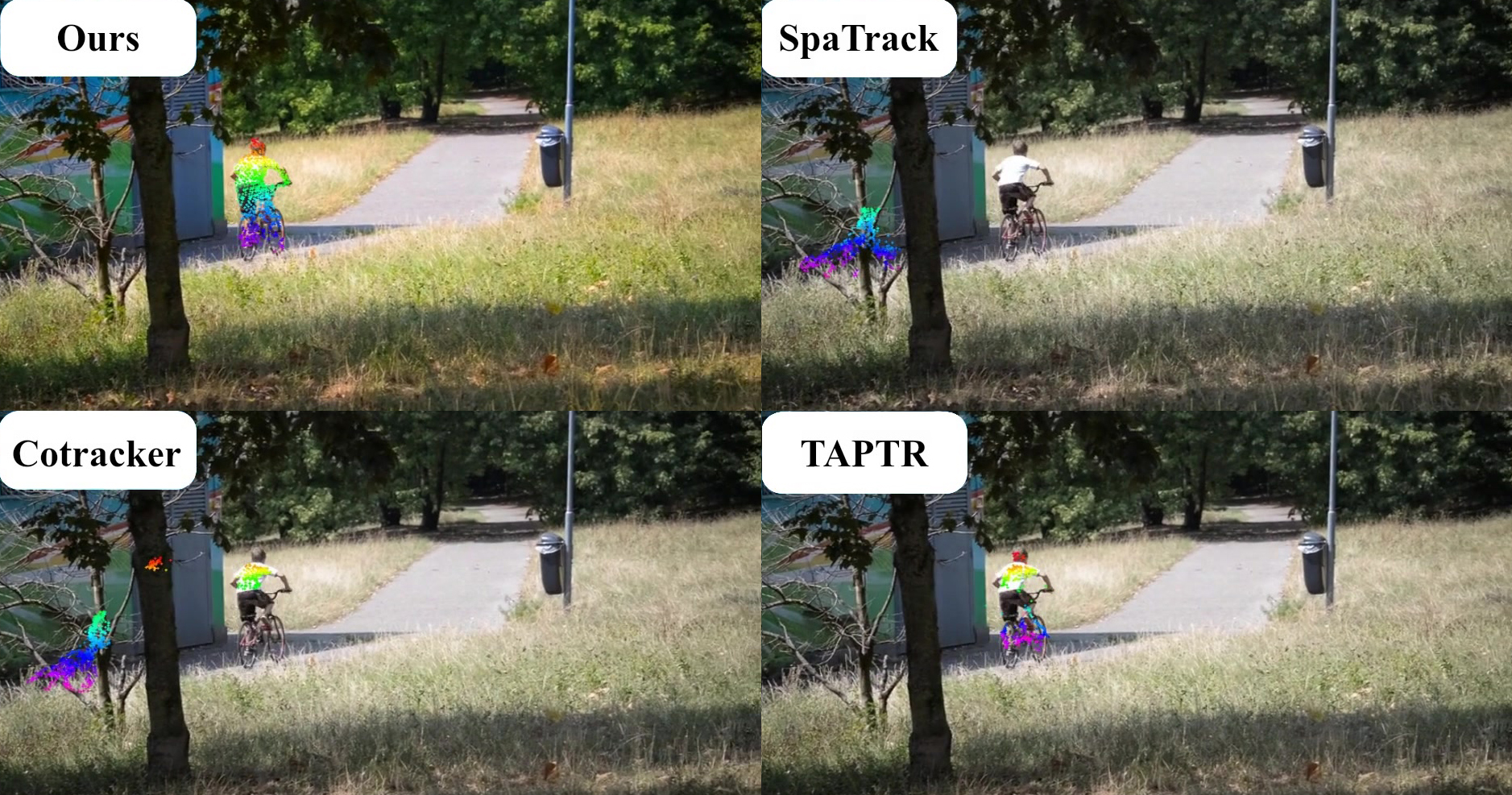} 
    \caption{Results of tracking at a higher frame rate. Sliding window based methods can easily lose track after occlusion and drift due to accumulating errors, while ours exhibit robustness.} 
    \label{fig:case3}  
\end{figure*}

\begin{figure*}[h]
    \centering
    \includegraphics[width=\linewidth]{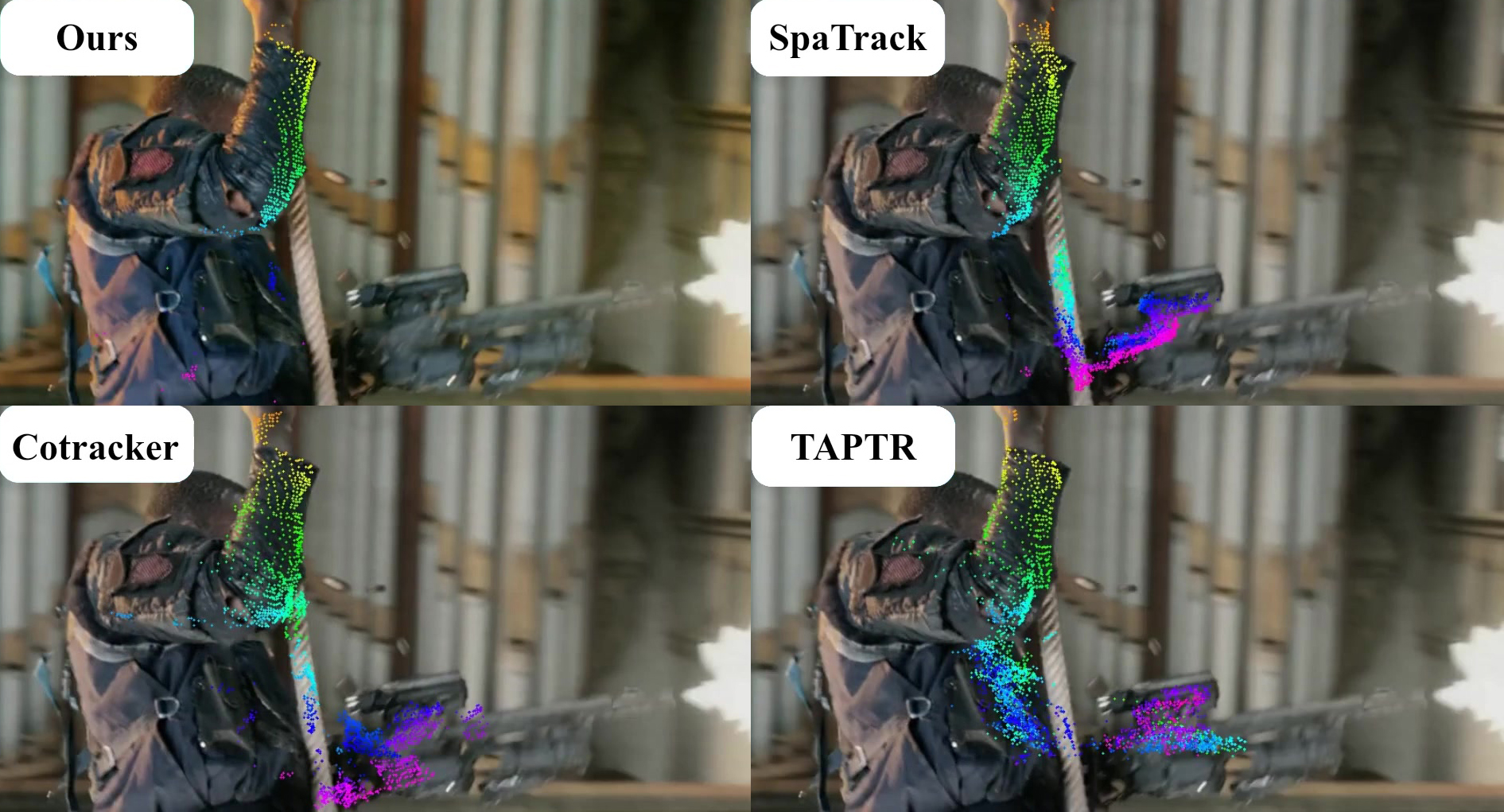} 
    \includegraphics[width=\linewidth]{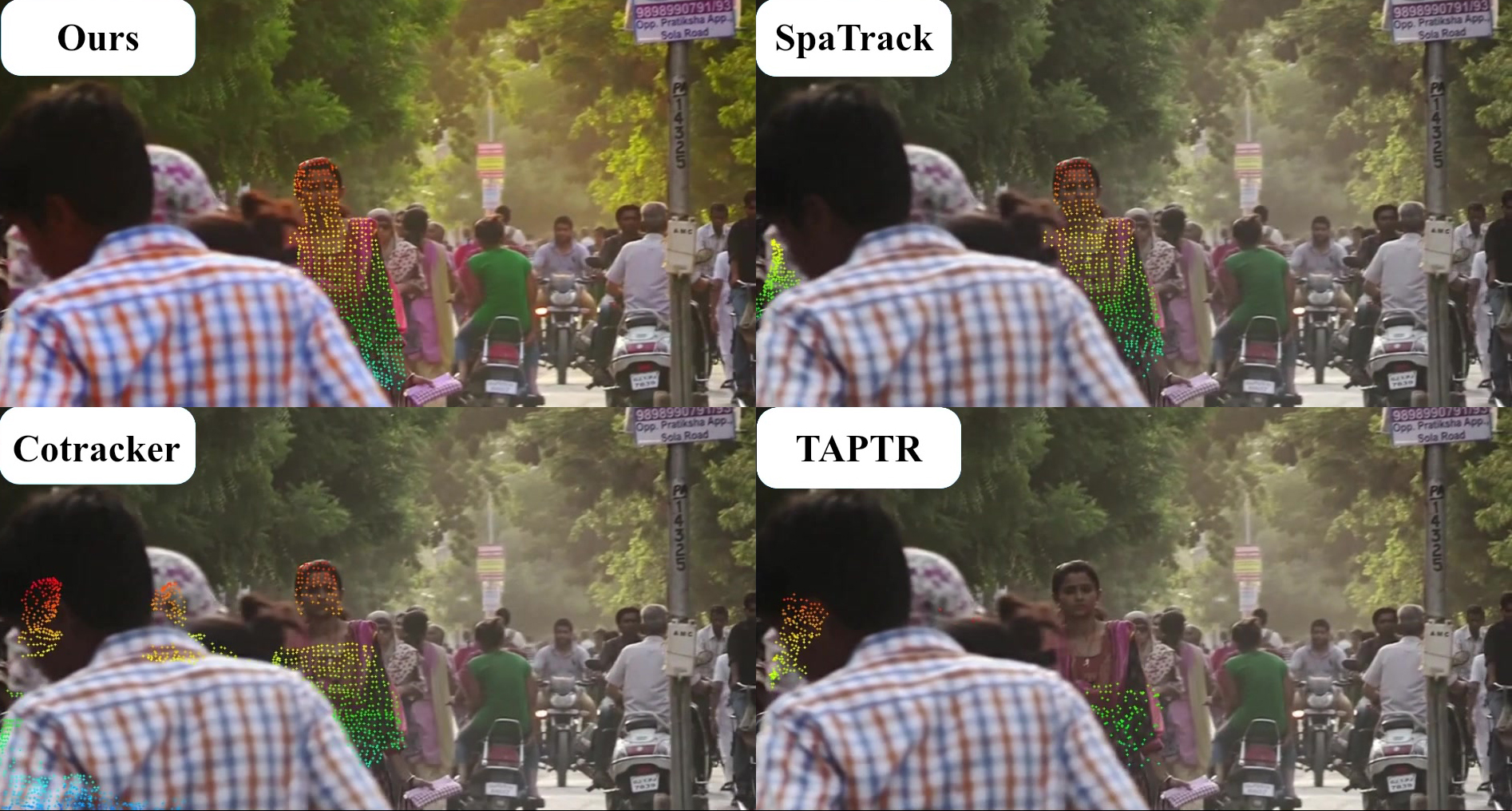} 
    \caption{Results of tracking at a higher frame rate. Sliding window based methods can mispredict points to other regions during occlusion (e.g. the gun and rope in \textit{shooting} and the wrong person in \textit{india}), while ours exhibit robustness.} 
    \label{fig:case4}  
\end{figure*}

To further demonstrate the robustness of our method, we conduct experiments on extended videos from TAP-Vid-DAVIS, simulating high frame-rate videos by repeating each frame three times, as illustrated in Fig.~\ref{fig:case3} and Fig.~\ref{fig:case4}. In contrast to typical sliding-window or flow-based trackers (such as TAPTR~\cite{taptr}, SpatialTracker~\cite{xiao2024spatialtracker} and Co-Tracker~\cite{karaev2023cotracker}), which tend to accumulate errors and drift over time, our integration of long-term key points with short-term optical flow enables continuous, drift-free tracking of the same point through occlusions.



\end{document}